\documentclass[11pt,a4paper]{article}
\usepackage[margin=2.5cm]{geometry}
\usepackage{natbib}

\usepackage{url,hyperref,microtype,subcaption,lineno}

\usepackage{graphicx}%
\usepackage{multirow}%
\usepackage{amsmath,amssymb,amsfonts}%
\usepackage{amsthm}%
\usepackage{mathrsfs}%
\usepackage[title]{appendix}%
\usepackage{xcolor}%
\usepackage{textcomp}%
\usepackage{manyfoot}%
\usepackage{booktabs}%
\usepackage{algorithm}%
\usepackage{algorithmicx}%
\usepackage{algpseudocode}%
\usepackage{listings}%
\usepackage[capitalize]{cleveref}
\usepackage{array}
\usepackage{diagbox}
\usepackage{float}
\usepackage{siunitx}
\usepackage{csquotes}
\usepackage{hyperref}
\usepackage{threeparttable}
\usepackage[onehalfspacing]{setspace}

\usepackage[draft]{changes}

\def\Authors{Raphael Memmesheimer\,$^{1,\dagger}$, Martina Overbeck\,$^{2,\dagger}$, Dominik Beyer\,$^{2,\dagger}$, Björn Kral\,$^{3}$,  Sabine Bellmann\,$^{2}$, Sven Schneider\,$^{4}$, Jan Zimmermann\,$^{4}$, Anna-Maria Meer\,$^{5}$, Medina Klicic\,$^{6}$, Simone Roth\,$^{6}$, Carolin Straßmann\,$^{6}$, Alexander Arntz\,$^{6}$, Marlene Wessels\,$^{7}$, Johannes Kraus\,$^{7}$, Paul Schweidler\,$^{8}$, Tristan Schnell\,$^{2}$, Christoph Zimmermann\,$^{2}$, Benedikt Pulver\,$^{9}$, Wilhelm Stork\,$^{2,10}$, Martin Gersch\,$^{3}$, Sven Behnke\,$^{1}$ and Arne Roennau\,$^{2,11}$}
\def\Address{
$^{1}$Autonomous Intelligent Systems, University of Bonn, Germany \\
$^{2}$FZI Forschungszentrum Informatik, Germany \\
$^{3}$School of Business and Economics, Freie Universität Berlin, Germany \\
$^{4}$Institute for Occupational Safety and Health of the German Social Accident Insurance (IFA), Germany \\
$^{5}$Fraunhofer Institute for Manufacturing Engineering and Automation IPA, Germany \\
$^{6}$Ruhr West University of Applied Sciences, Germany\\
$^{7}$Human Factors and Engineering Psychology, Johannes Gutenberg University Mainz, Germany\\
$^{8}$Human-Factors-Consult GmbH, Germany\\
$^{9}$TÜV SÜD Product Service GmbH, Germany \\
$^{10}$Institute for Information Processing Technologies, Karlsruhe Institute of Technology, Germany \\
$^{11}$Machine Intelligence and Robotics Lab, Karlsruhe Institute of Technology, Germany \\

$^{\dagger}$These authors contributed equally to this work.
}
\def\corrAuthor{Raphael Memmesheimer, Martina Overbeck, and Dominik Beyer}
\def\corrEmail{memmesheimer@ais.uni-bonn.de, overbeck@fzi.de, beyer@fzi.de}

\begin{document}
\onecolumn

\title{Benchmarking Robots for Everyday Environments: From Lab Experiments to Real-World Operations}


\author{\parbox{\linewidth}{\centering\normalsize \Authors}}
\date{}


\maketitle
\begin{center}\small\itshape \Address\end{center}
{\raggedright\noindent\textbf{Correspondence:} \corrAuthor\\
\texttt{\small \corrEmail}\par}
\vspace{1ex}

\begin{abstract}
This study presents an interdisciplinary process for benchmarking service robots in public everyday environments, addressing the gap between laboratory metrics and real-world operation at technology readiness levels 6--7. Three robots were evaluated in three use cases: outdoor park cleaning, pedestrian underpass cleaning, and interactive assistance in a public library. Over three years (2023--2025), seven benchmarking events (a consensus workshop and six on-site evaluations, two per use case) assessed the robots in increasingly realistic settings, from controlled conditions to public operation. An interdisciplinary panel of experts in robotics, human--robot interaction, safety, and economic viability consolidated use-case-specific metrics into a shared evaluation concept refined across three phases. Rather than ranking the robots, the evaluation examined how task fulfillment, interaction quality, safety, and economic viability interact in the transition from laboratory prototype to operational system. Direct quantitative comparability across the heterogeneous use cases remained limited, characteristic of benchmarking at this maturity level. The process yielded cross-case insights: increasing environmental realism introduces challenges that uniform metrics capture only in part; the interaction-quality guideline works best as a modular instrument adapted to each use case; safety was assessed by occupational-safety and product-certification experts who adapted industrial test procedures for public settings; and economic viability proved a cross-cutting dimension. We conclude that benchmarking at higher technology readiness levels is well served by a shared procedure for selecting and documenting metrics across these four categories, with safety as mandatory, rather than by a fixed set of metrics: generalizability lies in the procedure rather than in identical metrics.
\end{abstract}

\section{Introduction}\label{sec:intro}
Benchmarking is an important driver for innovation and progress in robotics. Especially when it comes to advancing robots from lab research to deployment in practice, the benchmarking of robots across comparable environments is a complex task that challenges existing methodology. In practice, deploying similar robots in everyday real-world scenarios requires the robots to perform simultaneously across multiple performance dimensions, such as task efficiency, safety, or interaction quality. Existing benchmarking efforts, particularly robotic competitions, commonly compare systems addressing the same predefined task in structured and reproducible settings and therefore provide limited guidance for heterogeneous applications with different purposes at high technology readiness levels (TRL). This paper addresses this gap by examining an interdisciplinary process for developing and refining use-case-specific metrics across different public-space robot applications and by deriving cross-case insights into trade-offs and limits of comparability under real-world conditions. To this end, we formed an interdisciplinary benchmarking panel of experts from robotics, human-robot interaction, safety, and economic viability to evaluate the performance of different robots over the course of three years in multiple public daily life scenarios under controlled evaluation conditions. The use cases and robots included were in libraries, parks, and pedestrian underpasses. As a first step, our approach involved defining custom metrics for each robot-use-case pair in the categories of task fulfillment, interaction quality, safety, and economic viability. They were then tested in three iterations while being refined and aligned with each other. The final goal was to achieve the interdisciplinary benchmarking of three TRL 6-7 everyday life scenarios. 

The contribution of this paper is threefold: (i) a \emph{process contribution}: an interdisciplinary, panel-based benchmarking methodology, inspired by the EuRoC freestyle format, in which use-case-specific metrics are proposed by the deployers of each robot and consolidated across disciplines in consensus meetings; (ii) a \emph{methodological contribution}: the resulting metric sets and evaluation instruments in the four categories of task fulfillment, interaction quality, safety, and economic viability for three public-space use cases, refined over three phases of increasing realism; and (iii) an \emph{empirical contribution}: cross-case findings from seven benchmarking events (a consensus workshop and six on-site evaluations) with three deployed TRL 6--7 robots, exposing which metrics, guidelines, and standards transfer across use cases and which do not.

The paper is structured as follows: In Section~\ref{sec:relatedwork}, we review related work on benchmarking frameworks. Section~\ref{sec:benchmarking_method} introduces our interdisciplinary benchmarking method, including the three-phase approach, four evaluation categories, three use cases, and the benchmarking panel. Section~\ref{sec:results} presents the evaluation concepts, metrics, and results from all three phases. Section~\ref{sec:lessons_learned} discusses lessons learned, and Section~\ref{sec:conclusion} concludes with contributions and future directions.

\section{Related Work}
\label{sec:relatedwork}

Before reviewing benchmarking approaches, we clarify the central terminology used throughout this paper. Following ISO 8373:2021~\citep{ISO8373}, a \emph{service robot} performs useful tasks for humans or equipment in personal or professional use; the standard explicitly lists guidance or information and cleaning as examples. Because no universally accepted definition of a \emph{social robot} exists \citep{henschel2021makes}, we adopt the established definition of \citet{bartneck2004design}, using the term for a physically embodied, autonomous or semi-autonomous robot that interacts and communicates with humans in accordance with expected behavioral norms. The term \emph{social service robot} denotes service robots whose task fulfillment inherently requires, or unavoidably entails, social interaction with humans in shared spaces. As the term is usually assigned by use case and task, not the robot per se, such robots in public or everyday environments are sometimes summarized as \textit{everyday robots}~\citep{Beyer2025}. The three benchmarked systems span this spectrum: the library robot (Pepper) is an interaction-centric social service robot, whereas the two cleaning robots (Angsa and Adlatus) are service robots whose deployment in public space nevertheless produces incidental social encounters with passersby. This is why interaction quality is benchmarked as a dedicated category for all three use cases. Nonetheless, this paper primarily uses the term \textit{service robot}.

\emph{Benchmarking} itself commonly denotes measuring a system's quality against an accepted standard~\citep{bonsignorio2015toward}. As no such standard yet exists for high-TRL service robots in everyday public environments (\cref{sec:relatedwork}), we use the term procedurally: metrics, test protocols, and pass/fail criteria are defined ex ante by an interdisciplinary panel, applied under controlled evaluation conditions, and refined across phases. Comparison thus occurs against these criteria, across phases, and, where admissible, across use cases; maturing such criteria toward accepted standards is part of this paper's contribution.

\subsection{Benchmarking Approaches for Task Fulfillment}
For the interdisciplinary benchmarking of robots in everyday life use cases with high TRL~\citep{ronnau2023towards}, related work is still scarce. A common way to benchmark rather uniform, task-based scenarios are robotic competitions~\citep{behnke2006robot,dias2016robot,nardi2016robotics}.

While the state of the art in (social) service robotics has seen a lot of progress recently~\citep{ghodke2024latest}, most robotics projects seem to be in the TRL 2-5 range~\citep{ronnau2023towards}.  Competitions and robotic challenges address this by providing benchmarks and boosting research~\citep{behnke2006robot}. Defining goals, procedures, and environments unknown to developers enables the comparison of different approaches and helps prepare technologies for use outside laboratories~\citep{ronnau2023towards,roa2021mobile}. Examples of such challenges are manifold. The \emph{DARPA Challenges} -- like the DARPA Grand Challenge 2005~\citep{thrun2006stanley} for autonomous vehicles, the DARPA Robotics Challenge 2015 for humanoid robots handling disaster use cases, and the DARPA Subterranean Challenge 2021~\citep{orekhov2022darpa} for cavern exploration—were competitions in the United States~\citep{nardi2016robotics}.
More recently, the \emph{ANA Avatar XPRIZE} pushed the development of immersive teleoperated avatar systems with a focus on haptics and interaction~\citep{BehnkeALRAM23,hauser2024analysisSORO,schwarz2023robust,lenz2025nimbro}; whereas the European Space Agency (ESA) co-organized the \emph{ESA-ESRIC Space Resources Challenges} with a focus on lunar exploration. The competition in the European Union, first, called for wheeled robots and mobile platforms to search for valuable resources on a simulated lunar surface (see e.g.,~\cite{ESAwebsite-first}, the winning approach~\cite{schnell2023efficient}) and, second, for lightweight rovers for excavation and beneficiation on the Moon~\citep{ESAwebsite, 2025DUST}.

Few competitions, however, tackle (public) daily life scenarios that resemble the use cases evaluated in this paper. There is, e.g., the \emph{RoboCup}, an international competition focusing on autonomous robots with ambitious long-term goals formulated for 2050~\citep{rossi2024human}. It is a platform for various applications and events, such as soccer~\citep{kitano1997robocup,GerndtSBSB:RAM15} and rescue~\citep{pellenz2016novel}, but also service robotics~\citep{wisspeintner2009robocup, matamoros2018robocup} in home environments. In \emph{RoboCup@Home}~\citep{wisspeintner2009robocup,memmesheimer2024robocup}, an interdisciplinary external jury evaluates the final performance by metrics regarding originality and presentation, relevance/usefulness to everyday life, and elegance/success of the overall demonstration. The latter was also represented in the \emph{European Robotics League Consumer Service Robots} (ERL Consumer)~\citep{studley2023perspective, basiri2019benchmarking}, where a benchmarking scenario for home applications and healthy aging, as well as independent living, was presented.  

Cleaning robotics also faces challenges addressed by competitions like the Automated Cleaning Challenge (Deutsche Bahn) and the Future Convenience Store Challenge (FCSC), which tested tasks such as cleaning train stations and customer toilets~\citep{DBCleaningChallenge2023,wada2017new}. While these competitions push robots beyond lab settings, they still rely on structured, reproducible scenarios to ensure fair comparison~\citep{ronnau2023towards}. However, they fall short of capturing real-world, high-TRL (6–7) conditions or benchmarking cross-scenario aspects, as they focus on predefined tasks in controlled environments~\citep{memmesheimer2024cleaning,RoboCupHomeRuleboook}.

A different approach is the EuRoC challenge, an EU project aimed at advancing European manufacturing~\citep{siciliano2014euroc}. Unlike traditional competitions, its Freestyle Stage allowed competitors to propose their own peer-reviewed goals, defining use cases and customized metrics for their robots~\citep{heppner2020fla2ir,awad2015european}. This flexibility accommodated diverse use cases, robots, and evaluation criteria.

Complementary to competition-based benchmarking, recent work documents robot deployments and encounters in public and semi-public settings. Field reports describe the practical and organizational challenges of running robots in public environments~\citep{bu2024field}, the behind-the-scenes work of Wizard-of-Oz operators controlling and troubleshooting robots in a public plaza~\citep{pelikan2025people}, and how passersby actually encounter, accommodate, and make sense of delivery robots~\citep{pelikan2024encountering}. Related studies derive design implications from the specific character of public space~\citep{pelikan2025making} and compare dyadic and group--robot interactions in a semi-public setting~\citep{mueller2024egocentric}. Our work complements this line of research on deployment practice, situated interaction, and public-space design by adding a structured, multi-category benchmarking process on top of comparable public-space deployments.

\subsection{Benchmarking Approaches for Interaction Quality}
\label{sec:relatedwork-interaction}



While task-oriented benchmarking evaluates the functional performance of robotic systems, the quality of human-robot interaction (HRI) is becoming increasingly important for everyday robots. Particularly in domestic, care, or public environments, a robot's success is determined not only by its ability to perform tasks correctly, but also by the manner in which it interacts with users~\citep{Mortezapour2025, Beyer2025}. Key benchmarks include, e.g., the interaction principles of ISO 9241-110:2020~\citep{ISO9241-110}, which address dimensions such as comprehensibility, predictability, trustworthiness, perceived presence, and acceptance~\citep{Coronado2022, Abrams2021}.
Recent work also considers perception-based metrics, for instance, related to a robot's audibility~\citep{agrawal2024sound, wessels2025auditory, cha2018effects, allen2025robots}.

Beyond momentary interaction quality, research on service robot adoption indicates that motivational, situational, and psychological factors, as well as trust-related differences across user segments, are associated with users' attitudes toward and intentions to use service robots in hospitality and entertainment contexts~\citep{binesh2023motivational,binesh2025unlocking,binesh2026user}. Trust beliefs and user dispositions similarly shape the acceptance of service robots across application contexts~\citep{kraus2024role,schuele2022patients}. These findings complement interaction-quality benchmarking with a broader user-acceptance perspective.
In evaluating interaction quality, user studies commonly employ standardized questionnaires to assess subjective evaluations of, e.g., perceived usability, trust in autonomous systems, or the social presence of the robot~\citep{Coronado2022, Mizuchi2020, babel2021investigating}. Additionally, behavior-based metrics are employed, including eye contact, response latencies, and the frequency of corrections or interventions~\citep{Mizuchi2020, Kompatsiari2019, Wiese2018}.

Unlike traditional performance metrics, interaction-related criteria are often subjective, context-dependent, and strongly influenced by individual user expectations~\citep{Murphy2013}. Valid measurement, therefore, usually requires a combination of several survey methods and cannot be achieved using individual, isolated metrics~\citep{Aly2017, Coronado2022}.

Early conceptual work emphasizes that interaction quality cannot be measured solely in functional terms but must also systematically consider psychological and social aspects of human perception~\citep{Kahn2007, Yanco2004}. More recent work builds on this approach by using continuously recorded signals, including physiological data. One example is the electroencephalogram (EEG)-based classification of problematic behavior in assistive robots, which was presented as part of an International Joint Conference on Artificial Intelligence (IJCAI) demo challenge~\citep{Chari2024}. A second example is the use of eye tracking in field settings to continuously assess users' attention in HRI~\citep{zeng2026encountering}.

A central line of research investigates the extent to which subjective interaction quality can be approximated by objective, observable factors~\citep{Yoshida2025, Mizuchi2023, Kokotinis2023, Mizuchi2020}.

Against this backdrop, recent HRI-centered workshops and competitions have highlighted the growing need for explicit interaction benchmarks, particularly in open and realistic scenarios~\citep{Tian2025, Hoggenmueller2025}. Examples include the 2024 HRI Robot Challenge Designing Social Robots in the Wild~\citep{HRI2024Website}, which addresses social interaction under real-world conditions; the Future Convenience Store Challenge, which focuses on customer interaction~\citep{kramer2021customer}; and the 2022 IROS Dialogue Robot Competition~\citep{Minato2023}, which evaluates dialogical skills, comprehensibility, and the appropriateness of human-robot communication. These competitions illustrate the shift from implicitly including interaction to treating it as an independent benchmark dimension, even though evaluations have thus far predominantly relied on human assessments.

Overall, interaction-related benchmarking complements task-oriented approaches by systematically integrating a human-centered perspective. This perspective is essential for robots used in everyday life because even technically capable systems can fail if they are perceived as incomprehensible, disruptive, or socially inappropriate~\citep{Goetz2003, Tian2021, Scheutz2011}.

\subsection{Benchmarking Approaches for Safety}
\label{sec:relatedwork-safety}


As the benchmarking process was conducted in the EU context, a range of applicable standards, regulations, and laws can be referenced as a baseline for safety-related requirements. However, these frameworks primarily support conformity assessment rather than providing direct benchmarking criteria for comparative evaluation of robotic systems.
Regulatory frameworks relevant in this context include the Machinery Regulation~\citep{EU2023mach}, the Electromagnetic Compatibility Directive 2014/30/EU~\citep{EU2014Electromagnetic}, the Radio Equipment Directive 2014/53/EU (RED,~\citep{EU2014RED}), the European Cyber Resilience Act (CRA)~\citep{EU2024CRA}, and the AI Act 2024/1689~\citep{EU2024AI_Act}.
These regulations are complemented by harmonised safety standards for robotic systems, many of which operationalise and further specify requirements of the Machinery Regulation, including ISO 12100:2010~\citep{ISO12100}, ISO 13849-1:2023~\citep{ISO13849_1}, EN 60204-1:2019~\citep{EN60204_1}, ISO 3691-4:2023~\citep{ISO3691_4}, ISO 10218-2:2025~\citep{ISO10218_2}, IEC 63327:2021~\citep{IEC63327}, and IEC 60335-1:2023~\citep{EN60335_1}, among others.


Mechanical hazards are among the most prominent in robots, even if others, such as electrical or thermal hazards, may also prevail. Safety focuses on diverting harm that may be caused by these hazards from persons. In robotics, many safety measures rely on technical solutions, the so-called safety functions, that range from sensory input, over the associated processing logic, to switching or controlling a machine's output. The safety functions’ reliability, or frequency of a dangerous failure, is represented by a Performance Level following the ISO 13849-1:2023~\citep{ISO13849} series of standards or a Safety Integrity Level (SIL) according to the IEC 61508:2010~\citep{IEC61508} series of standards. In either case, these quality criteria consider at least the severity of an injury, a person’s exposure time to the hazard, and a person’s possibility of avoiding the harm. Evaluating the safety of a machine or robot, for instance during a third-party conformity assessment, typically involves two phases: a theoretical review of design documentation or source code and practical experimentation on the system under test.

Several standards and other sources describe test pieces for evaluating a system’s detection capability regarding persons. IEC 61496-3:2018~\citep{IEC61496_3} for active opto-electronic protective devices responsive to diffuse reflection (AOPDDR) defines specific test pieces for both two-dimensional systems (AOPDDR-2D, such as planar safety laser scanners) and three-dimensional systems (AOPDDR-3D, such as multi-layered LIDARs). These test pieces are cylindrical or conical in shape to represent whole bodies, lower limbs, or body parts. They must be covered in black material with \qty{1.8}{\percent} remission that resembles black trousers fabricated from corduroy textile, white material with remission of \qtyrange{80}{90}{\percent}, and retro-reflective material that reflects and even focuses emitted light back onto the receiver.  The ISO 3691-4:2023~\citep{ISO3691_4} standard targets driverless industrial trucks and describes two cylindrical test pieces with \qtyrange{2}{6}{\percent} remission that represent a lying and a standing person. ISO 16001:2017 ~\citep{ISO16001} for earth-moving machinery compiles a list of test pieces for a wide range of sensor systems, including closed-circuit television (CCTV) cameras, various realizations of radar systems, or ultrasonic transceivers. 

The references above mostly apply to industrial settings. Hence, when benchmarking for everyday environments, they introduce implicit assumptions about the targeted persons. That is, they only apply to the working adult population that is healthy and has received an introduction to the risks associated with their workplaces. It also only includes harm targeting a single individual, not groups of people. All these assumptions may no longer hold in public spaces that are cohabited by vulnerable people, including children or the elderly, who are both unaware of the risks that a robot poses and unable to avoid imminent hazards. Especially in public spaces, it is additionally relevant to consider combined hazards that pose “indirect” risks to people. For example, a robot that collects trash could accidentally gather a person's medical supplies or push over a charcoal grill, thus scattering embers. Only a few standards include such challenges explicitly in their scope. Examples comprise the technical specification IEC/TS 62998-1:2019~\citep{IEC62998-1} for safety-related sensor systems and the ISO/IEC Guide 50:2014~\citep{ISOguide50}, which both contain information about considering children in safety. ISO 13482:2014~\citep{ISO13482} for personal care robots could be seen as a candidate for a safety standard in public spaces. 
The same applies to IEC 63327:2021~\citep{IEC63327} for commercial-use surface-cleaning robots, which at least acknowledges that such robots will work around large crowds of people.

Apart from the safety of hardware components and the robot in its entirety, aspects from other ethical, legal, or social perspectives (ELSI), which also include interaction quality, can have implications on the safety of a robotic system in public daily life scenarios as well. In Germany, operating a mobile robot in public environments invokes dealing, for example, with regulations on road traffic laws, depending on the exact use case~\citep{lehnsack2024}. These are therefore relevant when developing benchmarking processes for safety. In this context, safety refers to measures aimed at preventing harm to humans or the environment caused by robotic systems, whereas security-related aspects (e.g., protection of the robot against external interference or damage) are explicitly excluded in this work. It has to be noted that not all the mentioned safety aspects are fully evaluable within a single compressed benchmarking setup. 

\subsection{Benchmarking Approaches for Economic Feasibility}

Approaches to economic viability in technology-intensive and socio-technical domains commonly emphasize iterative, user-centered, and ecosystem-oriented methodologies as means to address uncertainty, complexity, and heterogeneous stakeholder settings~\citep{Bocken}. In service robotics, economic viability is rarely assessed as a static cost–benefit outcome but unfolds as a process in which value propositions, stakeholder roles, and resource configurations are aligned over time~\citep{Zott}. This dynamic is particularly pronounced in public-sector deployments, where economic considerations are tightly intertwined with organizational structures, public-sector logics, and institutional constraints~\citep{Carros}. Despite this conceptual breadth, the literature offers no established benchmarking format for the economic viability of service robots in everyday public environments.

Methodological contributions in this field frequently build on principles from Design Thinking and Lean Start-up, stressing phased processes of exploration, experimentation, validation, and refinement~\citep{Harms}. Rather than prescribing uniform evaluation schemes, the literature emphasizes the importance of selecting and adapting business model methods to specific use cases, maturity levels, and ecosystem constellations~\citep{Andreini}. Phase-specific process models are commonly proposed to structure this adaptation, distinguishing between early analytical and conceptual phases, subsequent validation of assumptions, and later stages of implementation~\citep{Frankenberger}. For interdisciplinary robotics consortia operating across heterogeneous use cases, this points toward consolidated, toolbox-based formats that combine methodological breadth with phase-specific guidance.

In line with these approaches, the economic benchmarking presented here is grounded in a toolbox-based framework consolidated by the transfer center for use by three competence centers~\citep{Kral}. The RimA Toolbox integrates established business model methods—including Value Proposition Canvas, Customer Journey, Stakeholder Network, and Business Model Canvas—within a coherent, phase-specific workflow and is continuously refined through application feedback from the competence centers. By combining structured guidance with iterative tool adaptation, this approach reflects prevailing perspectives on business model development~\citep{Foss} while extending them toward systematic economic benchmarking across different robot types, use cases, and ecosystem settings ~\citep{Bachmann}.

\section{Benchmarking Method}
\label{sec:benchmarking_method}

We aimed at benchmarking fundamentally different robot platforms in daily life scenarios, ranging from interaction-focused robots employed in libraries to cleaning robots in parks and pedestrian underpasses. This diverse set of platforms and use cases with different focuses guided us to a benchmarking approach inspired by the European Robotics Challenge approach~\citep{siciliano2014euroc}, where the participating teams could develop their own metrics in the different categories of \textit{task fulfillment, interaction quality, safety}, and \textit{economic viability}. 
In continuous exchange with a benchmarking panel, these metrics were refined and consolidated into an evaluation concept. This concept was subsequently applied across increasingly realistic phases, from conceptual development and laboratory evaluation to deployment in real-world environments.

\subsection{Benchmarking Phases}
\label{subsec:phases}
The benchmarking took place over a period of three years. Each year focused on an advancement from developing metrics to benchmarking under lab conditions and finally under practical conditions. In Phase 1, evaluation concepts, suggesting metrics for each of the categories (see \cref{sec:categories}) were developed by the deployers of each use case. In Phases 2 and 3, the metrics were applied under increasingly practical conditions. Phase 1 concluded with a consensus workshop; additional interim consensus meetings after Phases 1 and 2 incrementally developed the metrics in close discussion with the benchmarking panel. \cref{tab:phase_descriptions} gives an overview of all categories in relation to their phases.

\begin{table*}
    \centering

    \begin{tabular}{>{\raggedright\arraybackslash}p{4cm}|
    >{\raggedright\arraybackslash}p{.5cm}
    >{\raggedright\arraybackslash}p{3cm}    
    >{\raggedright\arraybackslash}p{3cm}
    >{\raggedright\arraybackslash}p{3cm}}
    \toprule
    \diagbox[width=8.4em]{\textbf{Category}}{\textbf{Phase}} & \multicolumn{2}{l}{\textbf{Phase 1}} & \textbf{Phase 2} & \textbf{Phase 3} \\
    \toprule
    \textbf{Task Fulfillment}  & \multirow{3}{*}{\rotatebox[origin=c]{90}{\parbox[c]{8cm}{\centering Use Case Selection}}} &  Justified selection and description of metrics and evaluation procedures & Evaluation in lab & Evaluation in practice \\
    \cmidrule (lr) {1-1} 
    \cmidrule (lr) {3-5} 
    \textbf{Interaction Quality} & & Justified selection and description of metrics and evaluation procedures & Pretests with self-selected participants & Field test with representative participants \\
    \cmidrule (lr) {1-1} 
      \cmidrule (lr) {3-5} 
    \textbf{Safety} & &Justified selection and description of criteria and verification procedures & Verification in lab & Validation during the field test \\
    \cmidrule (lr) {1-1} 
      \cmidrule (lr) {3-5} 
    \textbf{Economic Viability} & & Justified selection and description of tools and application procedures & Two to three (re-)selected tools applied & Business plan \\
    \midrule
     \textit{\textbf{Description}} & \multicolumn{2}{l}{\textit{Consensus workshop}} & \textit{Test in the presence of the benchmarking panel in own lab or in controlled settings} & \textit{Field tests in the presence of the benchmarking panel and presentation of project results at a central event} \\
    \bottomrule
    \end{tabular}
        \caption{Phase descriptions.}
    \label{tab:phase_descriptions}
\end{table*}

\subsection{Benchmarking Categories}
\label{sec:categories}
 
We benchmarked the robots in four categories that are aligned to reflect not only the technical performance of the robots but also the ELSI aspects. We therefore defined the categories of \textit{task fulfillment}, \textit{interaction}, \textit{safety}, and \textit{economic viability} to consider multiple dimensions for the benchmarking following~\cite{ronnau2023towards}. This is assuming that robots that operate in daily life should be useful, interact with the surrounding humans, and be safe. In an additional dimension, we addressed the economic viability supporting the deployers of each use case to develop and transition the use cases to be economically viable. Commonly, research groups concentrate on single dimensions, neglecting the existence of categories outside the subject's focus. The given categories were defined such that they are continuously present throughout the benchmarking. The given benchmarking approach is therefore highly interdisciplinary, designed to give equal importance to all the proposed benchmarking categories in different research disciplines. In the following, we present the benchmarking categories. 

\subsubsection{Task Fulfillment}

Task fulfillment measures the practical, quantifiable performance of the systems it was designed for. This category aims to establish objective metrics and standardized evaluation procedures to systematically assess performance. Typical metrics include success rates, completion times, and area processed per unit time. Since selecting these metrics was the initial phase of the benchmarking process, the chosen indicators will be presented in \cref{sec:results}. The category was overseen by experts with extensive experience in benchmarking robot systems outside laboratory settings, particularly through robotics research and robot competitions.

\subsubsection{Interaction Quality}

Interaction quality is also supposed to reflect the practical, measurable performance of the systems in interaction with a human user, even though it is more subjective in nature than task fulfillment (see \cref{sec:relatedwork-interaction}). The goal of the category is to develop metrics as objective and use-case universal as possible and evaluation procedures that systematically reflect the performance of each robot in a human-centered way. Common objective metrics are interaction principles of ISO 9241-110:2020~\citep{ISO9241-110} and the address of dimensions such as comprehensibility, predictability, trustworthiness, perceived presence, and acceptance~\citep{Coronado2022, Abrams2021}. As the choice of metrics was also the first phase of the benchmarking process, the chosen metrics will be displayed in \cref{sec:results}. The category was supervised by representatives with long-term experience in the benchmarking of robot systems out of lab conditions through HRI research.

\subsubsection{Safety}

To systematically assess safety in robotic systems, it is essential to define and justify key metrics—such as those related to mechanical, electrical, and functional safety—while outlining detailed verification procedures that align with specified performance levels.
The evaluation, therefore, aimed at multiple aspects, with a focus on mechanical safety (e.g., safety distances, potential impact forces during collaboration) and functional safety (e.g., performance level, response time).
For the safety evaluation, the evaluation concept was supposed to contain a detailed description of the verification procedure in relation to the required performance level. The safety evaluation was overseen by experts from an official German safety institution.

\subsubsection{Economic Viability}
\label{sssec:category_economic}

Unlike task fulfillment, interaction quality, and safety, economic viability followed a phase-specific evaluation logic. Consistent with the literature discussed above, it was assessed as a \emph{process}, using completion criteria tailored to each phase rather than uniform quantitative economic outcome measures; the corresponding results (\cref{ssec:results_economic}) therefore report process reflections rather than comparative measurements. Economic viability was addressed through a transfer-oriented three-phase process. RimA, the transfer center, provided the RimA Toolbox, a curated set of sixteen business model methods with associated tool guides, a process model, and templates, to the three competence centers, which applied and contextually adapted these methods within their respective use cases. Phase 1 required the selection and justification of four toolbox methods aligned with the specific characteristics of each service-robotics use case, consolidated through a consensus workshop. Phase 2 required the completed and documented application of two to three (re-)selected tools in real or near-real practice settings, including the testing of core assumptions through qualitative and quantitative feedback structured by a shared template with seven guiding questions. Phase 3 required the completion of a tailored business plan following a five-part template (value proposition, stakeholders and markets, financial structure, timeline, reflection on tool use), addressing either the micro level of the individual robotic application or the meso level of competence-center sustainability. Insights generated by the competence centers were systematically fed back into the Toolbox, enabling iterative refinement of methods and tool guides.

\subsection{Use Cases}
\label{ssec:use_cases}

\begin{figure*}
    \centering
    \includegraphics[height=.26\textwidth]{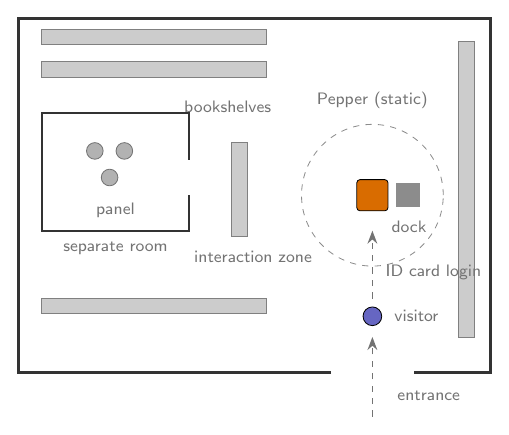}
    \hfill
    \includegraphics[height=.26\textwidth]{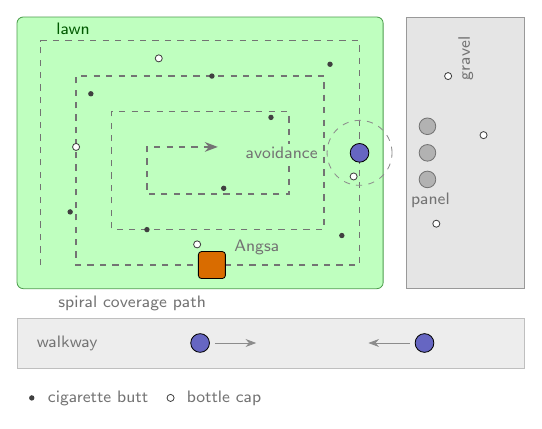}
    \hfill
    \includegraphics[height=.26\textwidth]{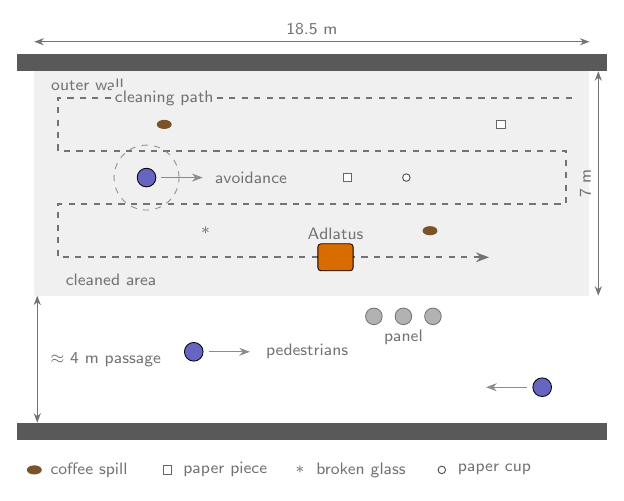}
    \caption{Abstract top-down schematics of the three benchmarking use cases: public library (left), park cleaning (center), and pedestrian underpass cleaning (right). Robots are shown as orange squares, humans as blue circles, infrastructure in gray, and motion or interaction zones as dashed lines.}
    \label{fig:use_case_schematics}
\end{figure*}

Three use cases were involved in the benchmarking attempt. These use cases deployed different robots: 
\begin{itemize}
    \item a Pepper robot~\citep{pandey2018mass} in a public library~\citep{helgert2024towards},
    \item an Angsa robot~\citep{angsawebsite} picking up small trash items in public parks~\citep{friedrich2025evaluating},
    \item and an Adlatus robot~\citep{adlatuswebsite} sweeping in a pedestrian underpass~\citep{raab2025assessingpedestrianbehaviorautonomous}.
\end{itemize}
In the following, we introduce the use cases underlying the benchmarking approach. An abstract, top-down overview of the three settings is depicted in \cref{fig:use_case_schematics}.

\begin{figure}
    \centering
    \includegraphics[width=0.8\linewidth]{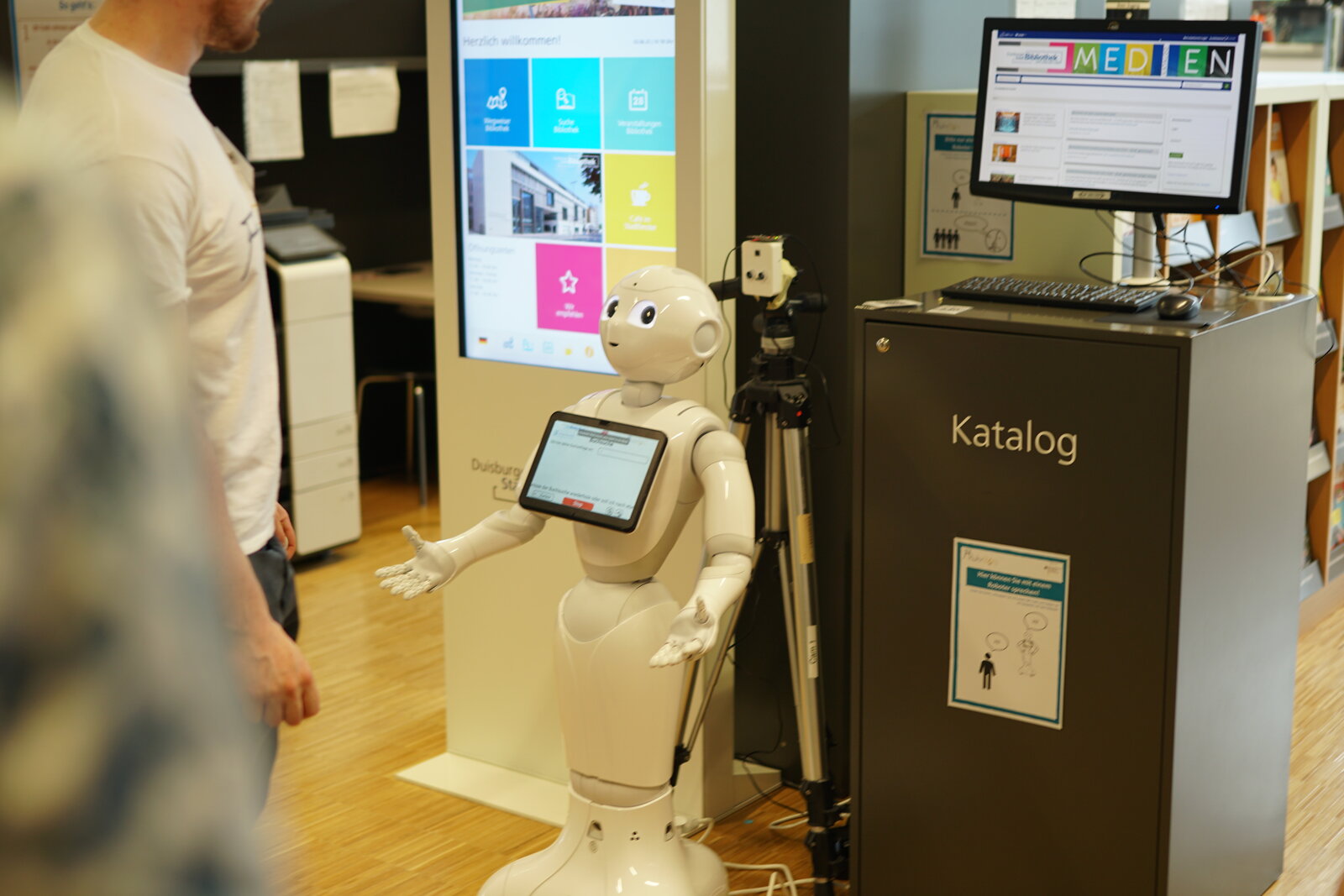}
    \caption{Public library use case: the Pepper robot providing book recommendations and reading samples to a visitor.}
    \label{fig:ruhrbots_teaser}
\end{figure}
\subsubsection{Public Libraries}

The aim of the robot interaction is to assist with book selection by providing book recommendations and reading samples, which the robot presents in the library~\citep{strassmann2024burgernahe}. This application goal was chosen based on focus groups and interviews conducted with citizens of the Ruhr region. 
For evaluation purposes, the goal was to engage as diverse a group of people as possible. This means including both regular library patrons and first-time visitors. Additionally, the focus was on reaching a wide range of user groups (e.g., in terms of age, gender, or German language proficiency) to test the diversity and inclusivity of the robot systems.
Participants receive a personalized or temporary robot ID card. The personalized card allows the robot to address participants by name during their first encounter. It also enables personalization for repeated measurements (returning participants) and recurring interactions with the robot, as well as appropriate language support if needed.
To begin the interaction, the participant holds their robot ID card in front of the Pepper robot~\citep{pandey2018mass}. Pepper is a 120 cm tall, 28 kg white humanoid robot with two arms, a torso, a single leg for standing, and a head. 
A representative image of the public library use case is depicted in~\cref{fig:ruhrbots_teaser}. The robot primarily communicates with users via speech and can display additional information on its tablet. 
The robot confirms the user login, for example, with a green checkmark on the screen, a brief welcome message, or an audible signal. The robot then greets the participant and briefly repeats the procedure.

\begin{figure}
    \centering
    \includegraphics[width=0.8\linewidth]{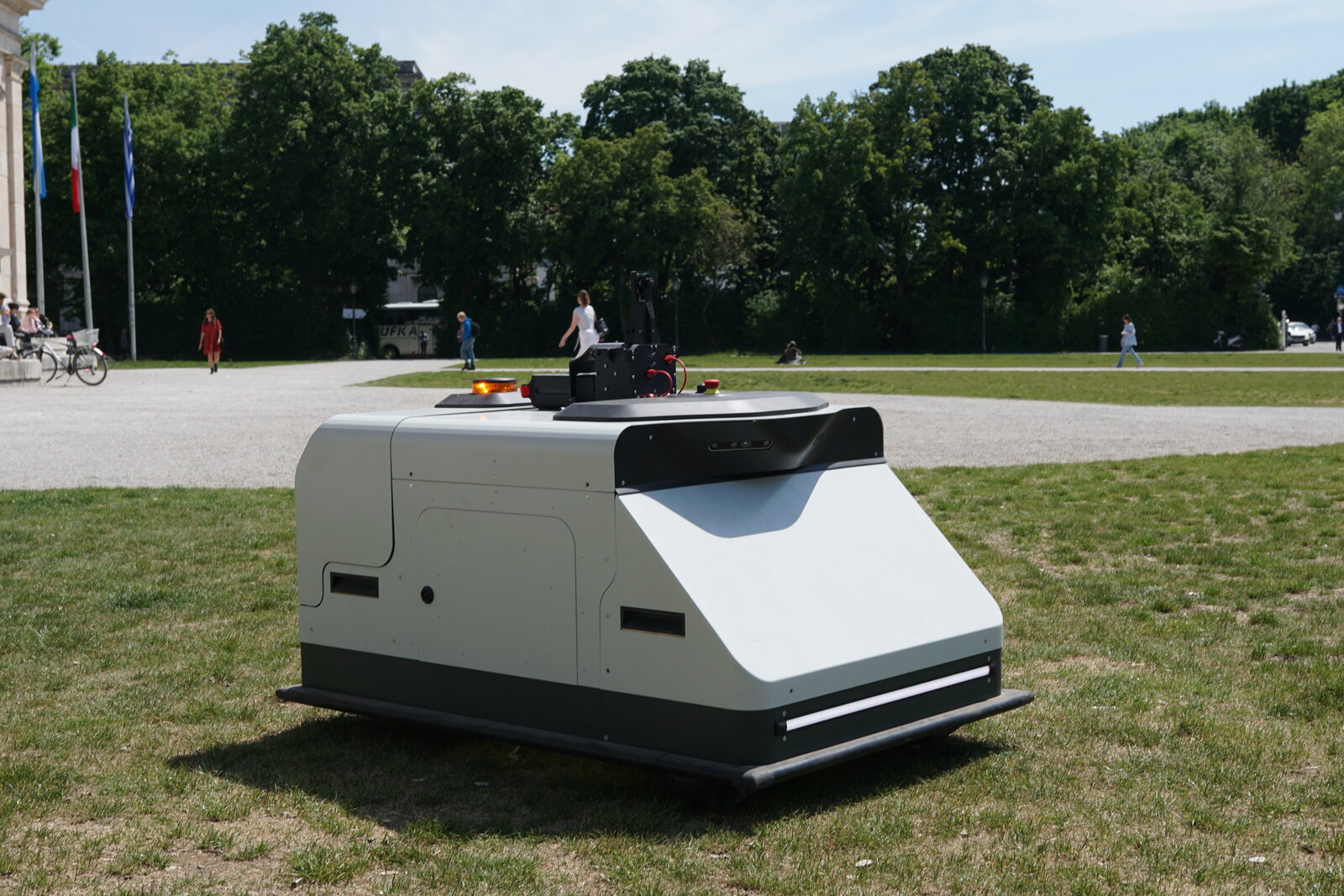}
    \caption{Park cleaning use case: the Angsa outdoor cleaning robot collecting small litter on a public lawn in Munich.}
    \label{fig:rokit_teaser}
\end{figure}
\subsubsection{Park Cleaning}


A mobile robot, approximately waist-high, moves at walking speed across a public lawn searching for small pieces of litter (cigarette butts and bottle caps)~\citep{tariq2024roboter}. An impression of the outdoor cleaning robot deployed in a public park in Munich is depicted in \cref{fig:rokit_teaser}.  As soon as a piece of litter is identified, the robot stops to suck it up. Since the robot moves in public spaces, encounters with people are to be expected. These encounters require interactivity. In the simplest case, this means that the robot stops before it comes into physical contact with persons to prevent injuries. Of course, it is desirable to have interactive capabilities that go beyond this level and also contribute to perceived safety, acceptance, user experience, and performance. The goal of interaction design is to minimize the impact of robot use on the important (social) functions of public space, such as recreation or sports, but also individual mobility. This applies not only to the immediate area of use but also to adjacent paths and roads. 

\subsubsection{Pedestrian Underpass Cleaning}

\begin{figure}
    \centering
    \includegraphics[width=0.8\linewidth]{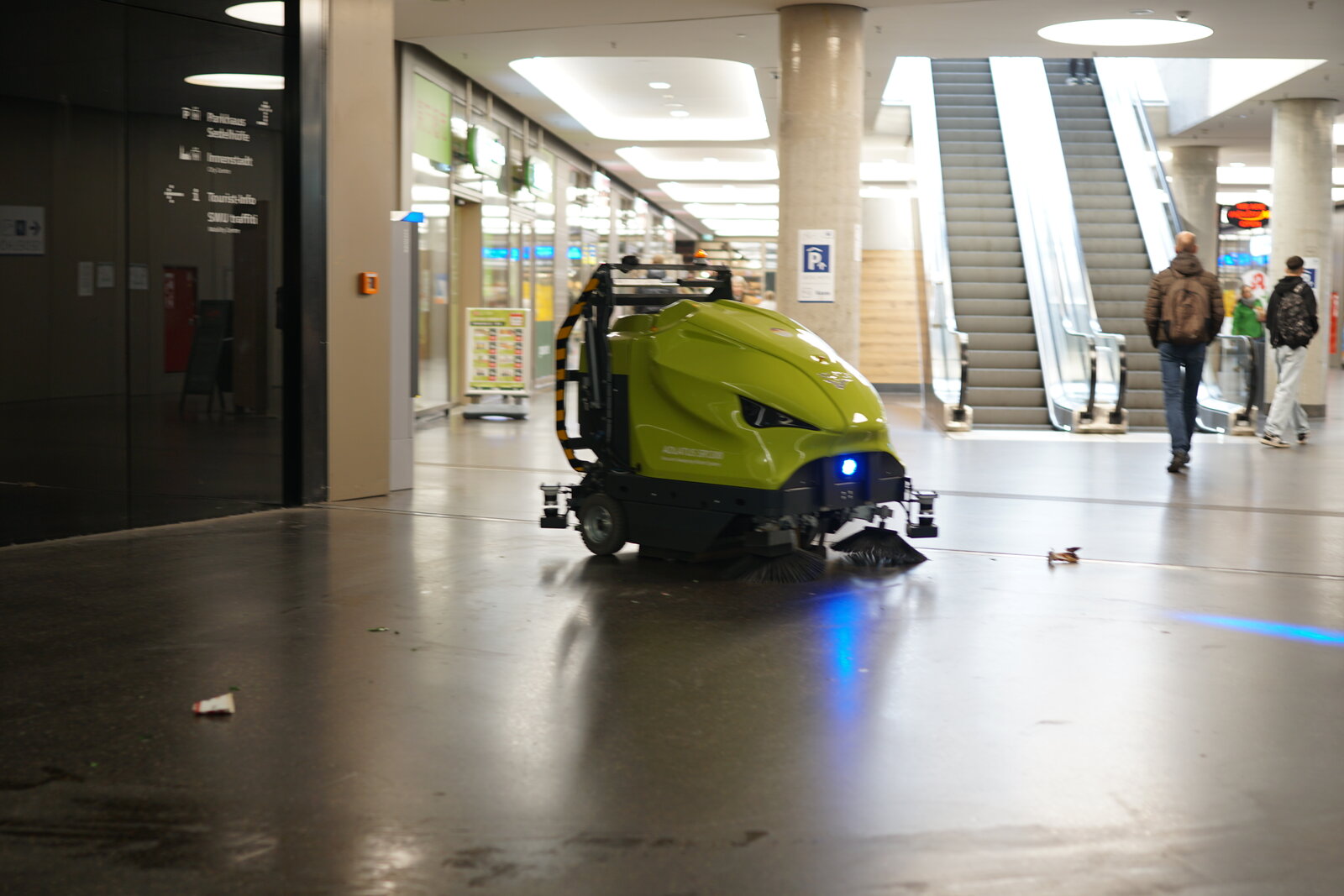}
    \caption{Pedestrian underpass use case: the Adlatus SR1300 sweeping robot cleaning the floor of a publicly accessible underpass at Ulm main station.}
    \label{fig:zenmri_teaser}
\end{figure}

In public spaces, robots operate among pedestrians without prior knowledge of robot behavior~\citep{kraus2024ulmer}. Impressions of the use case are depicted in \cref{fig:zenmri_teaser}.
In this use case, an Adlatus SR1300 sweeping robot cleans the floor in a publicly accessible underpass. The robot moves along a path parallel to an outer wall, maintaining a distance of about 2 meters. According to its cleaning plan, the robot turns on the spot at the end of its path and begins moving in the opposite direction. This creates a narrow passage, about 60 cm wide, between the robot and the wall. The challenge is to ensure navigation that is smooth, pleasant, and safe for all users while remaining efficient and uninterrupted for the robot. 

\subsection{Benchmarking Panel}
\label{ssec:benchmarking_panel}
An interdisciplinary benchmarking panel was formed to carry out the evaluations. As members, representatives from seven different institutions were included. These institutions were, on the one hand, the three project leads working on the chosen use cases, and on the other hand, experts on the benchmarking categories from the transfer project of the same German funding program. It, therefore, consisted of roboticists, psychologists specializing in HRI, safety experts, and business economists. 

The panel drew on a pool of representatives from these institutions. Attendance varied between locations, as individual members were substituted by colleagues from the same institution and discipline when they could not attend over the three-year period. Depending on the phase and test site, between four and seven panel members were present at each benchmarking event. In total, the panel consisted of 15 different individuals, one of whom attended all events and did a panel briefing before each evaluation to introduce the concept and ensure consistency.

\subsection{Evaluation Concept}
\label{ssec:evaluation_concept}
To initiate the interdisciplinary benchmarking process, the panel drafted an evaluation concept based on the four categories (Section~\ref{sec:categories}). In Phase 1, each robot's deployers defined use-case-specific metrics, which were then refined during a consensus workshop with the benchmarking panel. The panel's feedback was incorporated to finalize the evaluation concept.

In Phase 2, the practical evaluation was executed under lab conditions. For each use case, the benchmarking panel came together for a one-day evaluation of the system in an environment of their choice.
Phase 3 evaluated the use cases in practice in realistic environments. 

\subsection{Benchmarking Procedure and Materials} 
\label{ssec:procedure_materials}

To support the repeatability of the benchmarking events, this subsection summarizes the test procedures, materials, and system configurations per use case, as well as the way in which the observations of the panel members were converted into the reported results. The complete evaluation sheets, checklists, and the interaction-quality guideline for all three use cases are provided in the Supplementary Material.

\emph{Public library.} A SoftBank Pepper robot with a large language model-based dialog backend was deployed statically, attached to its charging station; no navigation was performed. Interactions were initiated by holding a robot ID card in front of the robot and followed the dialog structure described in \cref{ssec:use_cases}. Each interaction was observed by the panel and scored with a structured evaluation sheet containing the checklist items of \cref{par:res_phase1_ruhrbots}. Each checklist item was rated on a 0--5 point scale, the category scores were weighted by their importance for the book recommendation interaction (introduction \SI{10}{\percent}, book recommendation \SI{60}{\percent}, farewell \SI{10}{\percent}, system functionality \SI{20}{\percent}), and the weighted sum was normalized to a percentage, yielding the reported dialog success rate.

\emph{Park cleaning.} An Angsa outdoor cleaning robot operated in autonomous mode at a target speed of \SI{0.3}{\metre\per\second}, covering the designated area in a spiral pattern that circles from the outer boundary of the area toward its center (cf.\ \cref{fig:use_case_schematics}). The test areas (\SI{56}{\metre\squared} synthetic lawn on concrete in Phase 2; a designated lawn area and a gravel subarea in front of the Glyptothek in Munich in Phase 3) were prepared by panel members who distributed a counted set of litter items (cigarette butts and bottle caps) according to a standardized distribution procedure. Hits and misses were counted against this known ground truth after each run and documented photographically (before/after pictures).

\emph{Pedestrian underpass cleaning.} An Adlatus SR1300 sweeping robot followed its stored cleaning plan at its standard working speed, covering the defined area in a meander pattern with lanes parallel to the outer wall and turns on the spot at the lane ends, starting from the rightmost corner of the area (cf.\ \cref{fig:use_case_schematics}). The test areas (\SI{6.5}{\metre} $\times$ \SI{10.4}{\metre} in Phase 2; \SI{7}{\metre} $\times$ \SI{18.5}{\metre} in Phase 3) were prepared with counted impurities (paper pieces, coffee spills, and in Phase 3 additionally broken glass and paper cups) and static obstacles (persons, luggage, bins) at predefined positions. Cleaning completeness was verified by manual inspection and photographic documentation; efficiency was computed from the measured area and cleaning duration.

For interaction quality, each attending panel member individually completed the respective questionnaire or guideline (10-point rating scales per category plus qualitative comment fields); ratings were discussed and consolidated in a debriefing at the end of each event. For safety, each test (e.g., emergency stop, obstacle detection with test pieces, drop test) was assessed with a predefined pass/fail criterion by the safety experts of the panel. As the Phase 3 events took place in genuinely public environments, parameters such as weather, ambient noise, network quality, and visitor traffic could not be controlled; they were documented at each event and are reported alongside the results where relevant. The robots themselves were continuously developed configurations of commercial platforms; exact software version identifiers were not systematically recorded during the three-year process and can therefore not be reported retrospectively, which we acknowledge as a limitation for exact reproducibility.

\section{Results}
\label{sec:results}
The three phases were executed as planned with the benchmarking panel in attendance (see \cref{tab:phase_exec}). Following the development of the evaluation concepts for the three use cases, a consensus workshop meeting was conducted in Bonn in December 2023 (Phase 1). The more structured, secured benchmarking in Phase 2 happened in 2024 and 2025 in Bottrop, Stuttgart, and Ulm. Finally, also in 2024 and 2025, Phase 3 took place in Ulm, Duisburg, and Munich. In total, the three-year process thus comprised seven benchmarking events: the Phase 1 consensus workshop and six on-site evaluations --- two per use case, one each under Phase 2 and Phase 3 conditions. Interim consensus meetings after Phases 1 and 2 supported the refinement but are not counted. Because each evaluation was a one-day panel event, the per-event sample sizes are inherently small; they are therefore reported explicitly alongside each result below, and their implications are discussed in \cref{sec:lessons_learned}.

\begin{table*}

    \centering
    \begin{tabular}{>{\raggedright\arraybackslash}p{4cm}|
    >{\raggedright\arraybackslash}p{.5cm}
    >{\raggedright\arraybackslash}p{3cm}    
    >{\raggedright\arraybackslash}p{3cm}
    >{\raggedright\arraybackslash}p{3cm}}
    \toprule
    \diagbox[width=8.4em]{\textbf{Use Case}}{\textbf{Phase}} & \multicolumn{2}{l}{\textbf{Phase 1}} & \textbf{Phase 2} & \textbf{Phase 3} \\
    \toprule
    \textbf{Pedestrian Underpass Use case}  & \multirow{3}{*}{\rotatebox[origin=c]{90}{\parbox[c]{4cm}{\centering Evaluation Concept}}} &  Consensus Meeting in Bonn & Secured area in underpass in Ulm & Underpass in Ulm \\
    \cmidrule (lr) {1-1} 
    \cmidrule (lr) {3-5} 
    \textbf{Library use case} & & Consensus Meeting in Bonn & Laboratory in Bottrop & Field test in public library in Duisburg \\
    \cmidrule (lr) {1-1} 
      \cmidrule (lr) {3-5} 
    \textbf{Park use case} & &Consensus Meeting in Bonn & Controlled area with synthetic lawn in Stuttgart & Public lawn area in Munich \\
    \bottomrule
    \end{tabular}
        \caption{Phase execution. The seven benchmarking events took place as follows: consensus workshop in Bonn (December 2023); on-site evaluations for Public Library (November 2024; June 2025), Park Cleaning (November 2024; May 2025), and Pedestrian Underpass Cleaning (November 2024; June 2025).}
    \label{tab:phase_exec}
\end{table*}

\subsection{Task Fulfillment Results}

In the following, we present the results of the use cases regarding the task fulfillment, categorized into the different phases. We first present the derived metrics from the evaluation concept and then present the results from Phase 2 experiments under lab conditions through Phase 3 under public deployment conditions.

\subsubsection{Results from Phase 1 (Task Fulfillment)}

\subparagraph*{Results from Public Library (Task Fulfillment, Phase 1)}
\label{par:res_phase1_ruhrbots}

In the \emph{Public Library} use case, the task fulfillment was focused on the interaction with the user; therefore, two metrics, \emph{Dialog Success—Book Recommendation} and \emph{Dialog Success—Reading Aloud Phase} were derived. Both metrics were based on measuring the success using a sequential structured checklist. 
The checklist for a successful book recommendation dialog with Pepper includes:
\begin{itemize}
 \item Greeting \& Introduction: Pepper greets the user and explains its functions.
 \item Genre Selection: Pepper asks for the preferred book genre (e.g., crime, thriller, sci-fi, romance) and can list available genres.
 \item Recommendation Process: Pepper provides book suggestions (title, author, summary) and asks if the user wants more recommendations or to add a book to a wishlist.
 \item Wishlist Management: Pepper can list wishlist items and provide additional details (publication year, publisher, location, audiobook availability, ISBN, series).
  \item Voice Adaptation: Pepper adjusts speech (speed, pitch, volume) and can repeat or cancel its last output.
\end{itemize}
and for the successful reading session includes:
\begin{itemize}
    \item Greeting: Pepper greets the user if not done during the recommendation phase.
    \item Book Selection: Pepper asks which book from the wishlist the user wants to hear, and can list available options.
    \item Reading Preferences: Pepper asks if the user wants a short or long reading sample and can read the blurb, short   excerpt, or long excerpt.
    \item Continuation \& Adaptation: Pepper asks if the user wants to hear more books from the list and can adjust its voice (speed, pitch, volume).
    \item Natural Interaction: The dialog must be informative, truthful, relevant, clear, adaptable, and allow for corrections. Pepper can repeat, cancel, or \textquote{think} (verbally/non-verbally) as needed.
    \item Farewell: Pepper says goodbye to the user.
\end{itemize}

\subparagraph*{Results from Park Cleaning (Task Fulfillment, Phase 1)}

For the park cleaning use case, metrics regarding the \emph{Reliability of Waste Removal} and \emph{Efficiency of Waste Removal} were proposed, and after consensus, an agreement was reached with the benchmarking panel. 
The \emph{Reliability of Waste Removal} is evaluated by how reliably the system detects small litter items (e.g., bottle caps, cigarette butts) and distinguishes them from other objects. Performance is assessed using a 4-field table (hit, miss, false positive, false negative), with specificity and sensitivity as critical metrics. Tests are conducted on two different surfaces (e.g., grass and cobblestone), with standardized litter distribution.
The \emph{Efficiency of Waste Removal} is measured by the area cleaned per hour. The time required for cleaning is recorded, and results are adjusted for cleaning quality to ensure meaningful comparison.

\subparagraph*{Results from Pedestrian Underpass Cleaning (Task Fulfillment, Phase 1)}

For the pedestrian underpass cleaning use case, the following metrics were derived \emph{Completeness and Quality} and \emph{Efficiency}. These metrics are closely related to the previous use case. For the \emph{Completeness and Quality} the robot must thoroughly clean areas of varying sizes, including corners and edges, and handle different surface types and levels of contamination. Testing is conducted on a $20 \times 60$ meter area featuring three obstacles (a person, a trash bin, and a suitcase) and three types of litter (liquid spills, paper waste, and broken glass). The quality of cleaning is verified through manual inspection and photographic documentation.

The \emph{Efficiency} is measured by the robot’s ability to clean the test area from the previous one within a reasonable time. Key metrics include cleaning speed (\SI{}{\meter\squared\per\minute}), total cleaning duration, and the need for emptying the robot’s waste containers. Performance is evaluated based on the area cleaned per minute.

\subsubsection{Results from Phase 2 (Task Fulfillment)}

\begin{figure*}
    \centering
    \includegraphics[width=.32\textwidth]{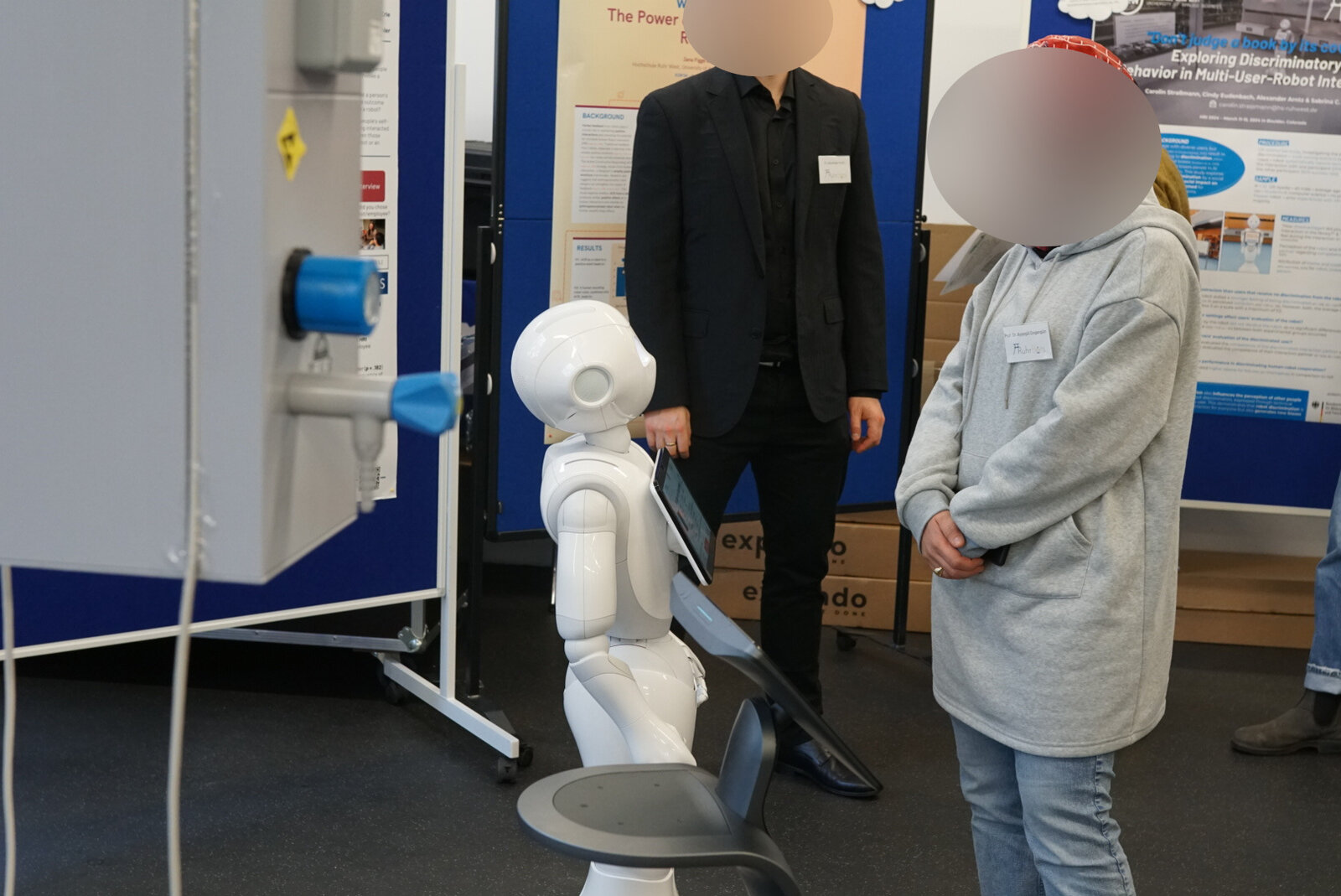}
    \hfill
    \includegraphics[width=.32\textwidth]{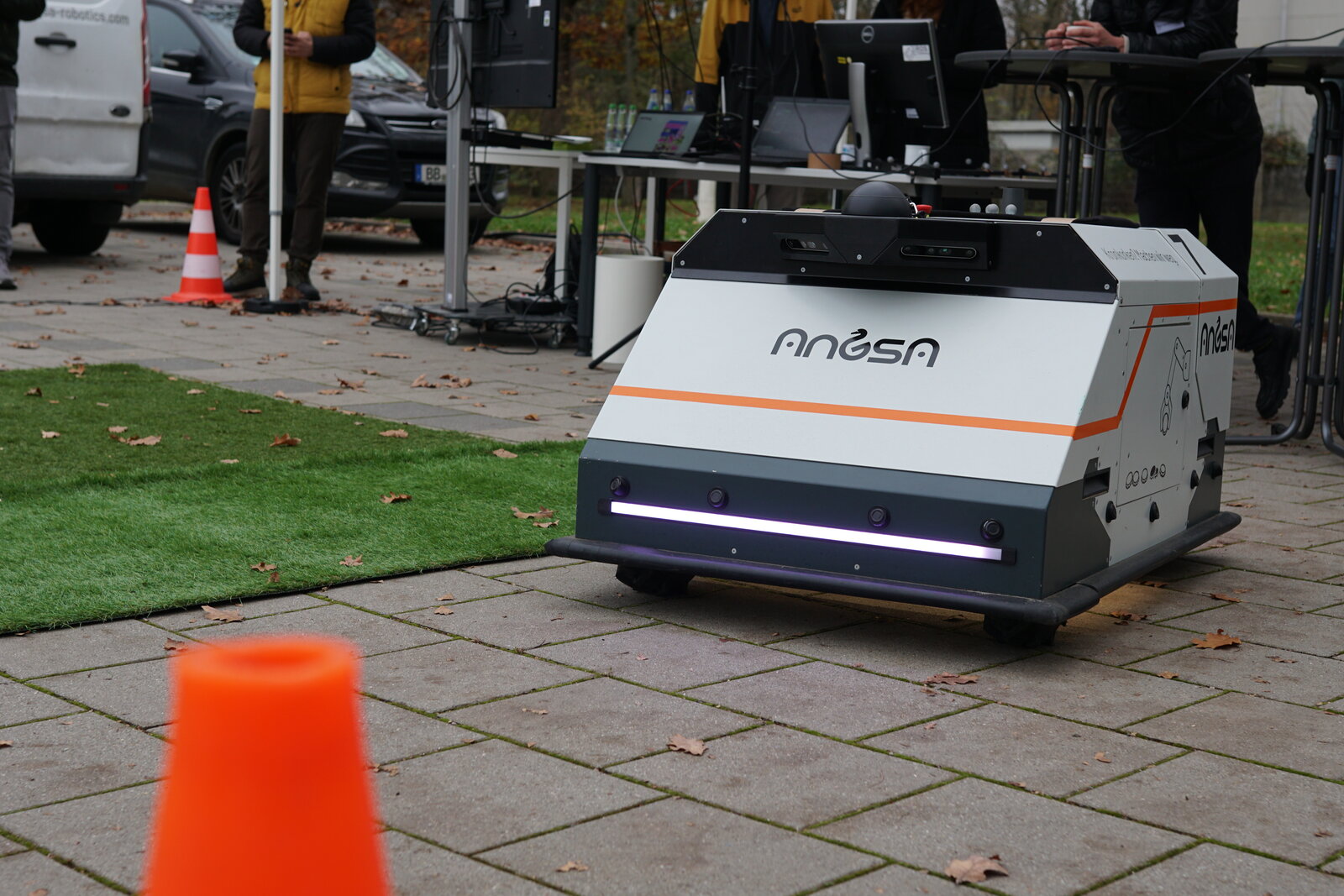}
    \hfill
    \includegraphics[width=.32\textwidth]{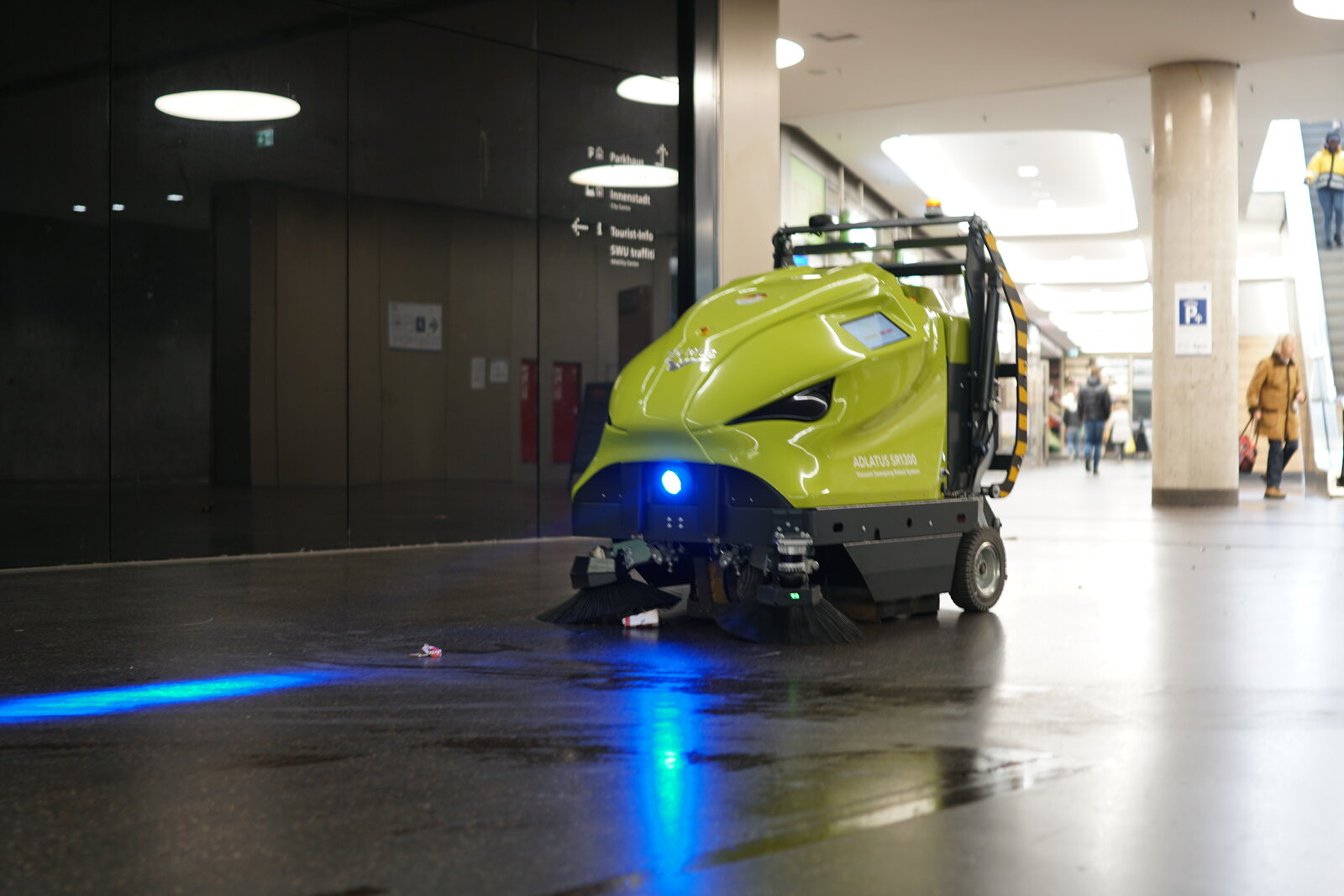}
    \caption{Exemplary impressions from the Phase 2 evaluations across the three use cases: the public library (left), park cleaning (center), and pedestrian underpass cleaning (right).}
    \label{fig:phase_2_overview}
\end{figure*}

Impressions from the Phase 2 evaluations are depicted in \cref{fig:phase_2_overview}. It follows a description of the results from the Phase 2 evaluations per use case.

\subparagraph*{Results from Public Library (Task Fulfillment, Phase 2)}
\label{sec:results-taskfulfil-library_phase2}

\begin{figure}[h]

    \centering

    \includegraphics[height=.32\linewidth]{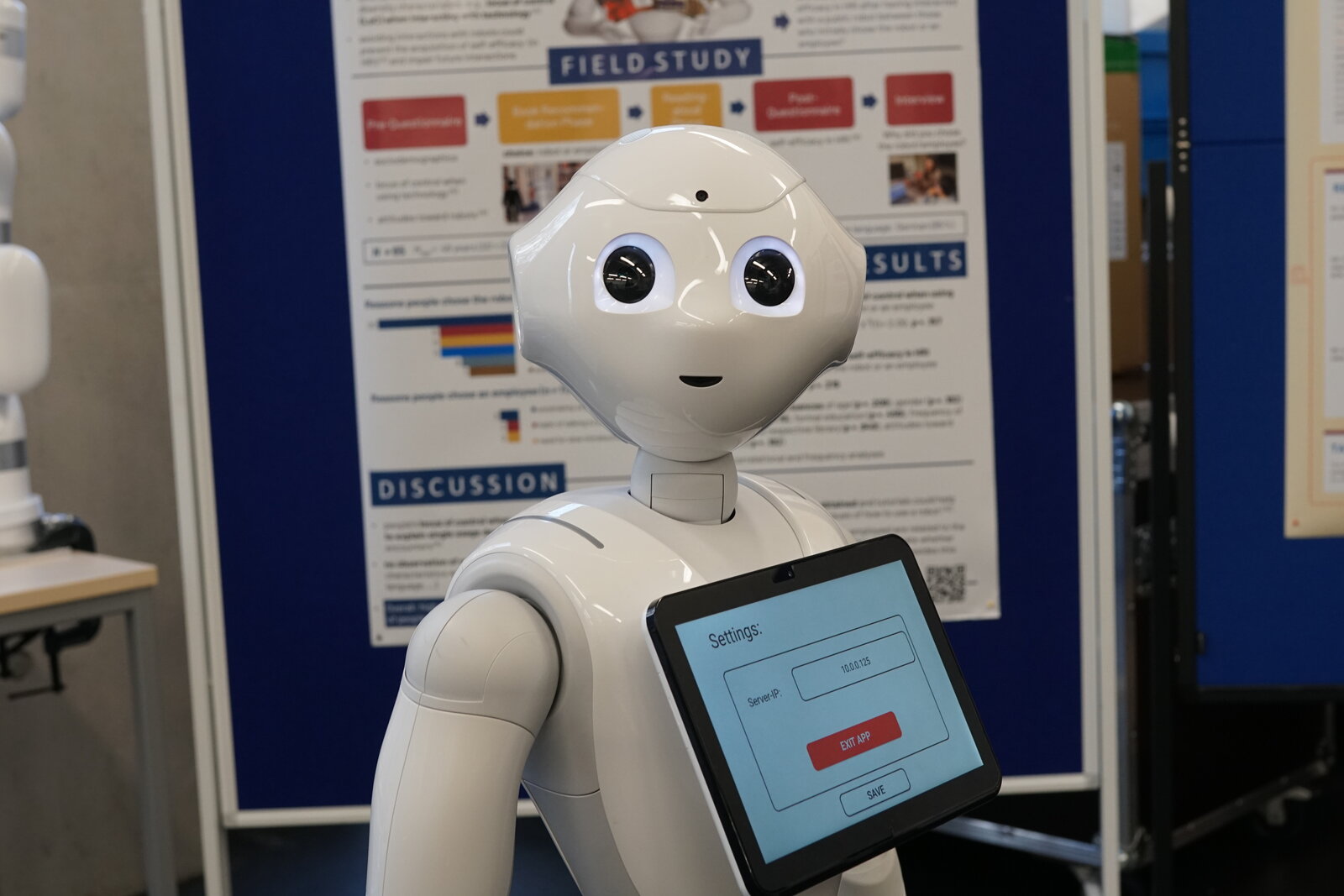}
    \hfill
    \includegraphics[height=.32\linewidth]{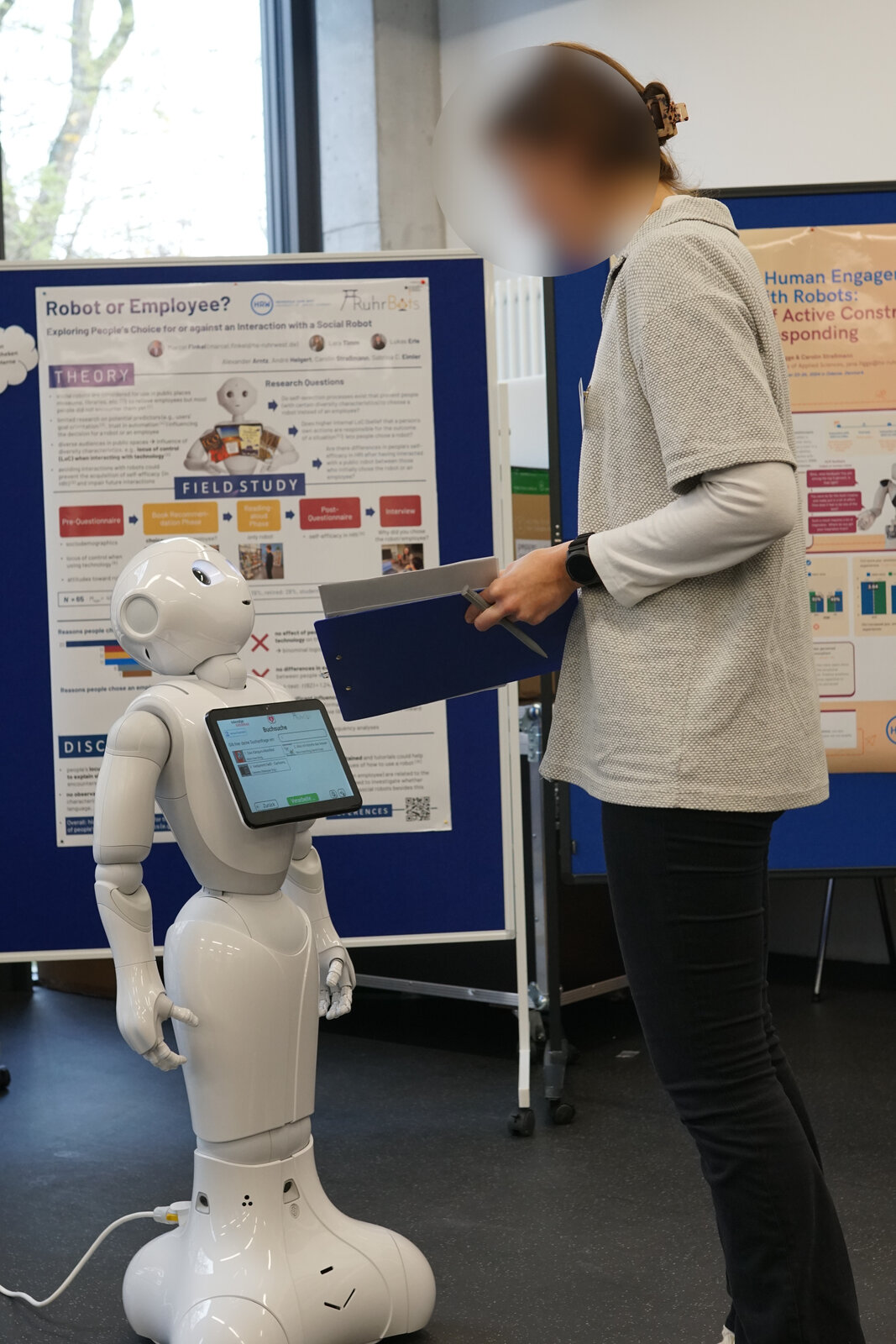}
    \hfill
    \includegraphics[height=.32\linewidth]{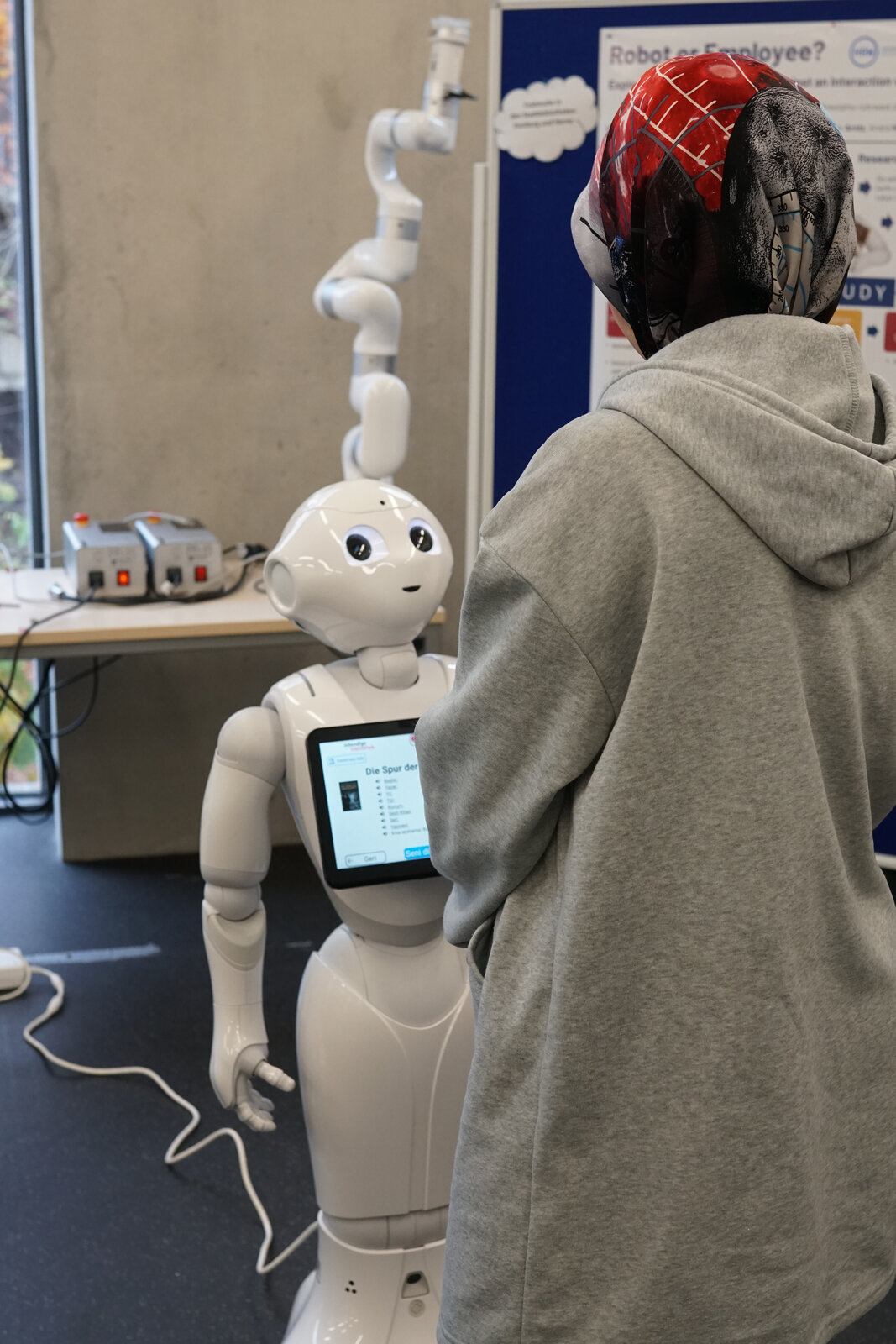}    
    \caption{Exemplary impressions from the public library use case under lab conditions during the Phase 2 evaluation with the Pepper robot waiting for a user to approach (left) and within an interaction (center \& right).}
    \label{fig:example_ruhrbots_phase2}
\end{figure}

The \textit{Public Library} use case in Phase 2 was evaluated in a lab room at the Bottrop campus of the University of Applied Sciences Ruhr West (see \cref{fig:phase_2_overview,fig:example_ruhrbots_phase2}).
The task fulfillment evaluation focused on book recommendations by natural interaction. No navigation to the final book recommendations was intended. For this use case, a user study evaluation sheet was constructed and exemplarily applied by observing the interaction between a user and the robot. 
The evaluation focused on the SoftBank Pepper robot. During Phase 2 testing, the multilingual capabilities were tested by employing individuals with different language backgrounds (German, English, and Turkish). The \emph{Dialog Success} was evaluated using an evaluation sheet that contained the proposed metrics from \cref{par:res_phase1_ruhrbots} weighted by their importance for the book recommendation interaction (see \cref{ssec:procedure_materials} for the weighting and scoring procedure). The dialog success rate evaluation yielded a result of \SI{35.7}{\percent} (based on one systematically scored interaction; further informal interactions informed the qualitative observations). The specific failures regarding the interaction originated from missing functionality that could not be triggered (adding and listing the items on the watchlist) and unintuitive feedback by the robot during the interaction. 


\subparagraph*{Results from Park Cleaning (Task Fulfillment, Phase 2)}

\begin{table} 
    \centering

    \begin{tabular}{c|c|c|c|c|c}
        \toprule
         \textbf{Experiment} & \textbf{Duration} & \textbf{Total} & \textbf{Cigarettes} & \textbf{Bottle Caps} & \textbf{Miss}\\
         \midrule
         1 & 8:32 & 24 &  14 & 10 & 7\\
         2 & 6:37 & 18 &  10 & 8 & 7 \\ 
         \bottomrule
    \end{tabular}
    \caption{Park Cleaning Task Fulfillment Results for Phase 2}
    \label{tab:results_rokit_phase2}
\end{table}

\begin{figure}
    \centering
      \includegraphics[width=.3\linewidth]{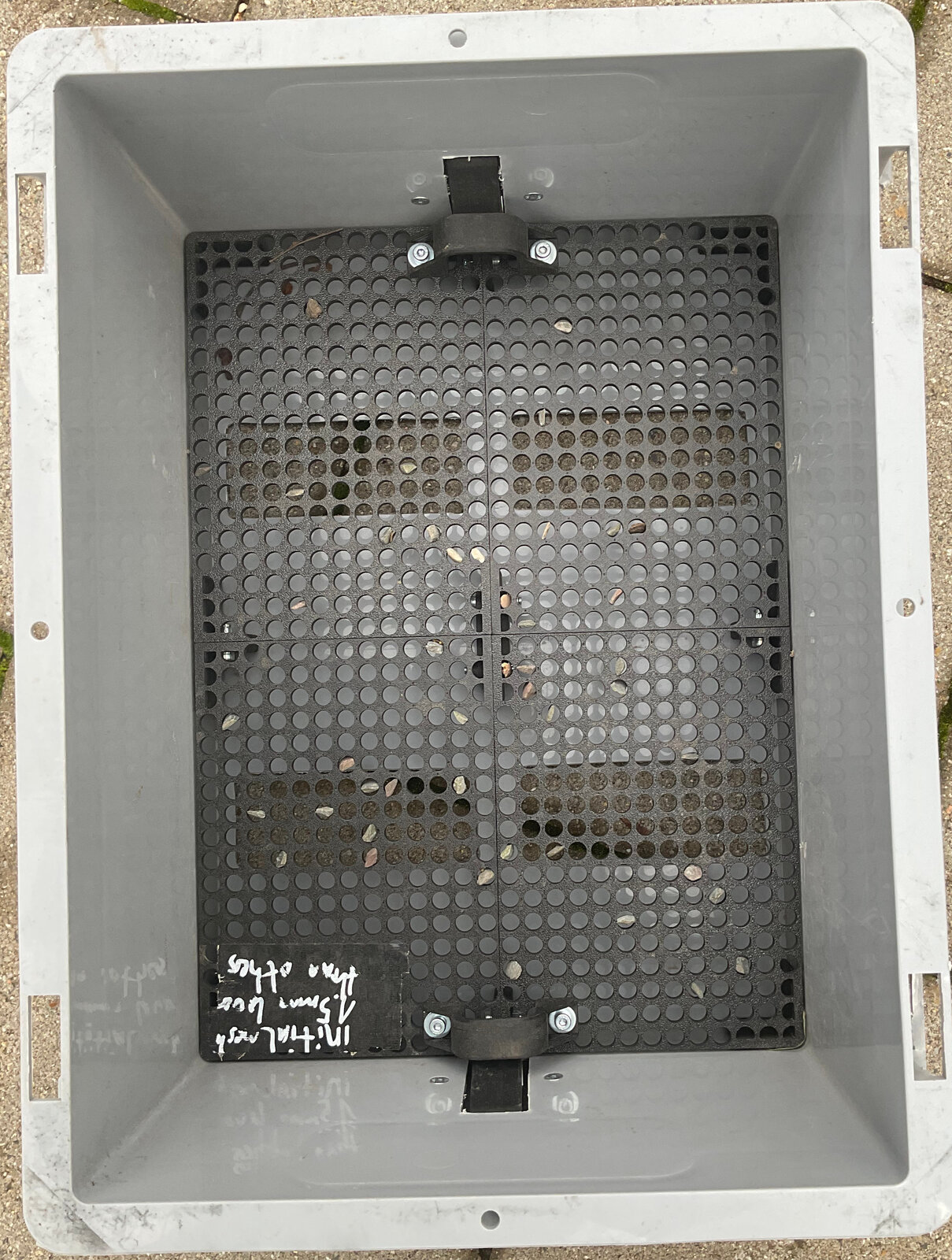}
    \hfill
      \includegraphics[width=.3\linewidth]{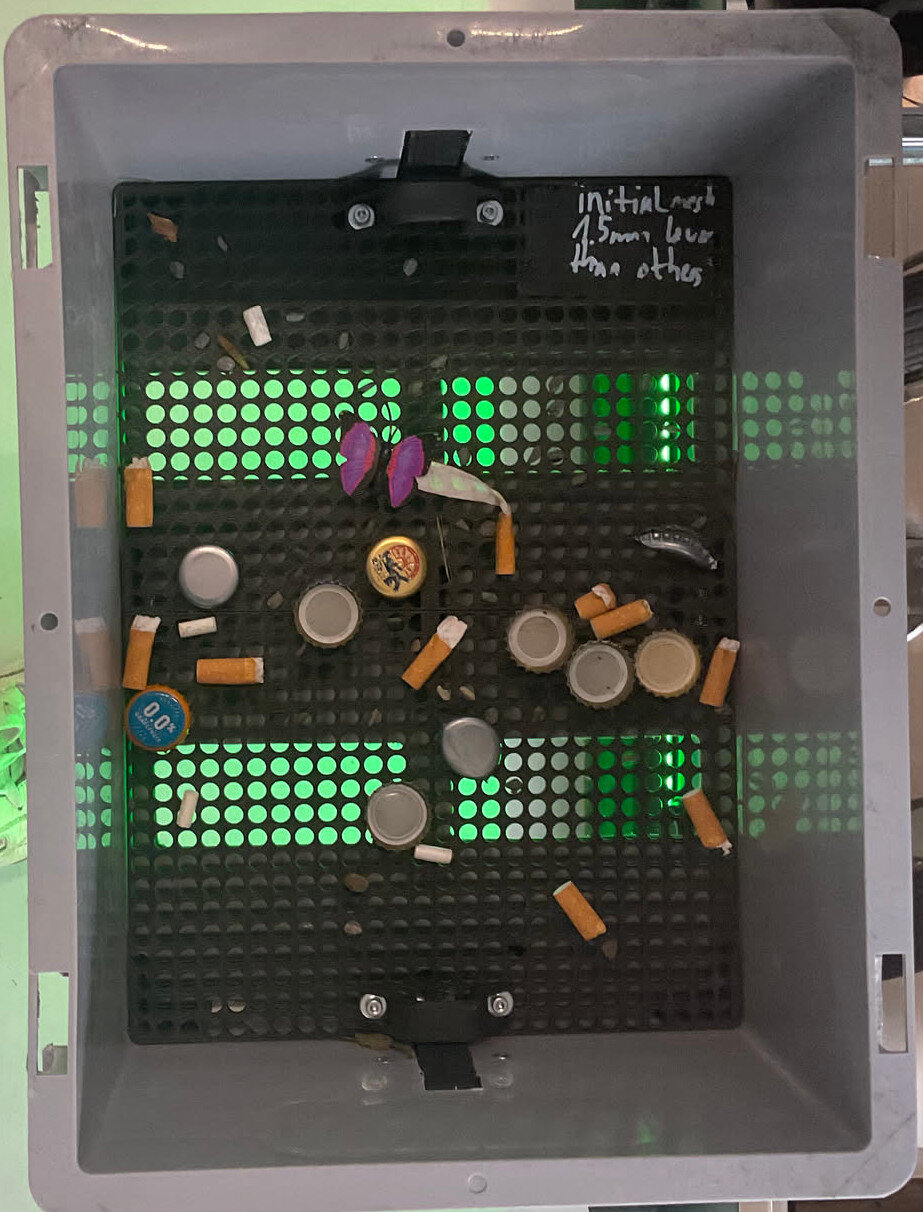}
      \hfill
      \includegraphics[width=.3\linewidth]{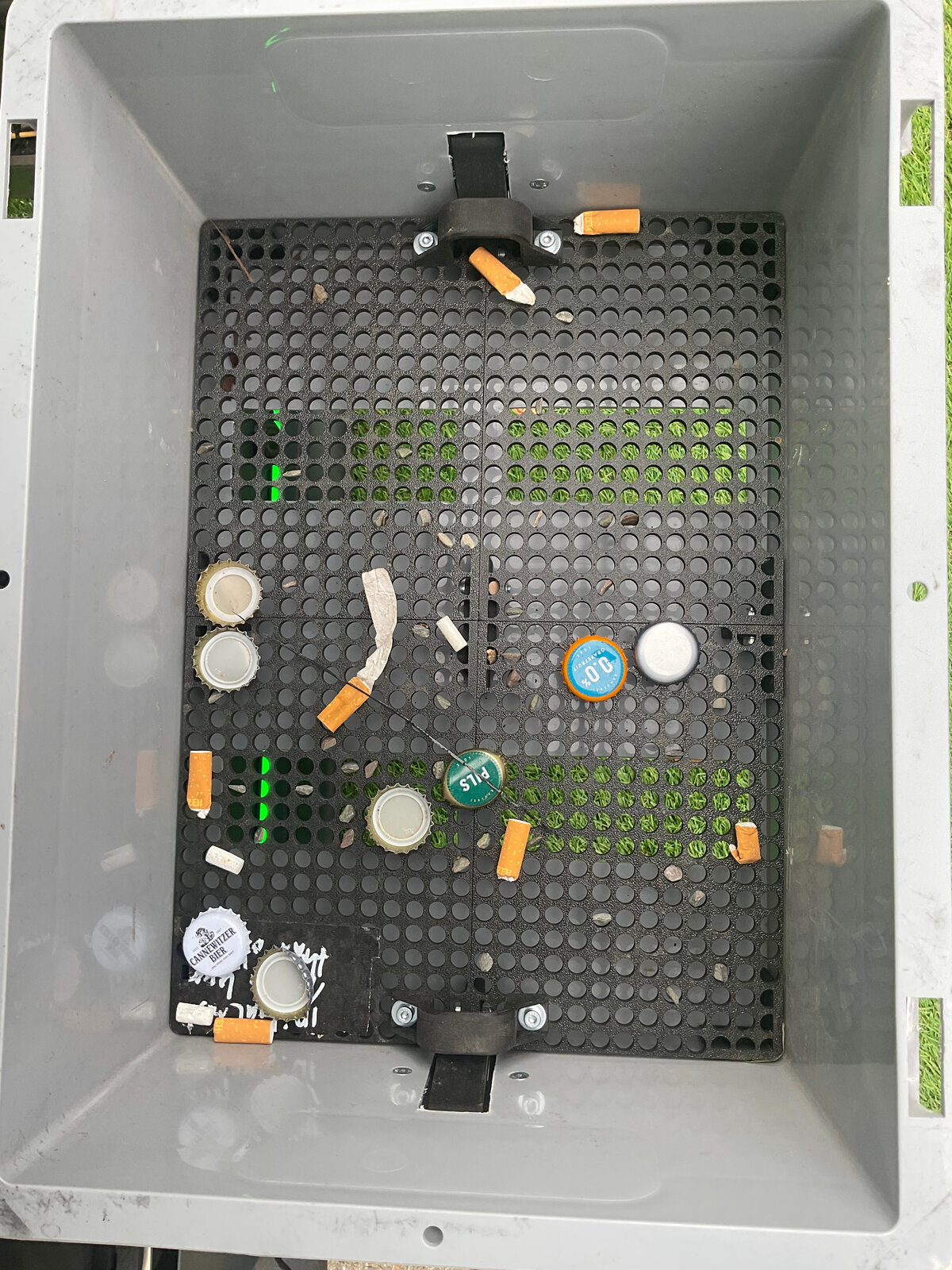}
    \caption{Before (left) and after Run 1 (center) and Run 2 (right) samples for park cleaning task fulfillment evaluation.}
    \label{fig:before_after_rokit_phase_2}
\end{figure}

The park cleaning use case was evaluated using an Angsa outdoor cleaning robot in front of Arena 2036~\citep{dittmann2019forschungscampus} in Stuttgart (see \cref{fig:phase_2_overview}). A dedicated area of \SI{56}{\metre\squared} to clean was constructed on a concrete ground. A square of synthetic lawn mimicked a meadow area. Trash in the form of cigarettes and bottle caps was distributed around the entire area. In total, the park cleaning use case was evaluated for two runs ($N=2$). The results are given in \cref{tab:results_rokit_phase2}. The missed trash items were mostly out of distribution, e.g., rubber bands or cable ties. Before and after pictures are shown in \cref{fig:before_after_rokit_phase_2}.

\subparagraph*{Results from Pedestrian Underpass Cleaning (Task Fulfillment, Phase 2)}

\begin{figure} 

    \centering

    \includegraphics[height=.22\linewidth]{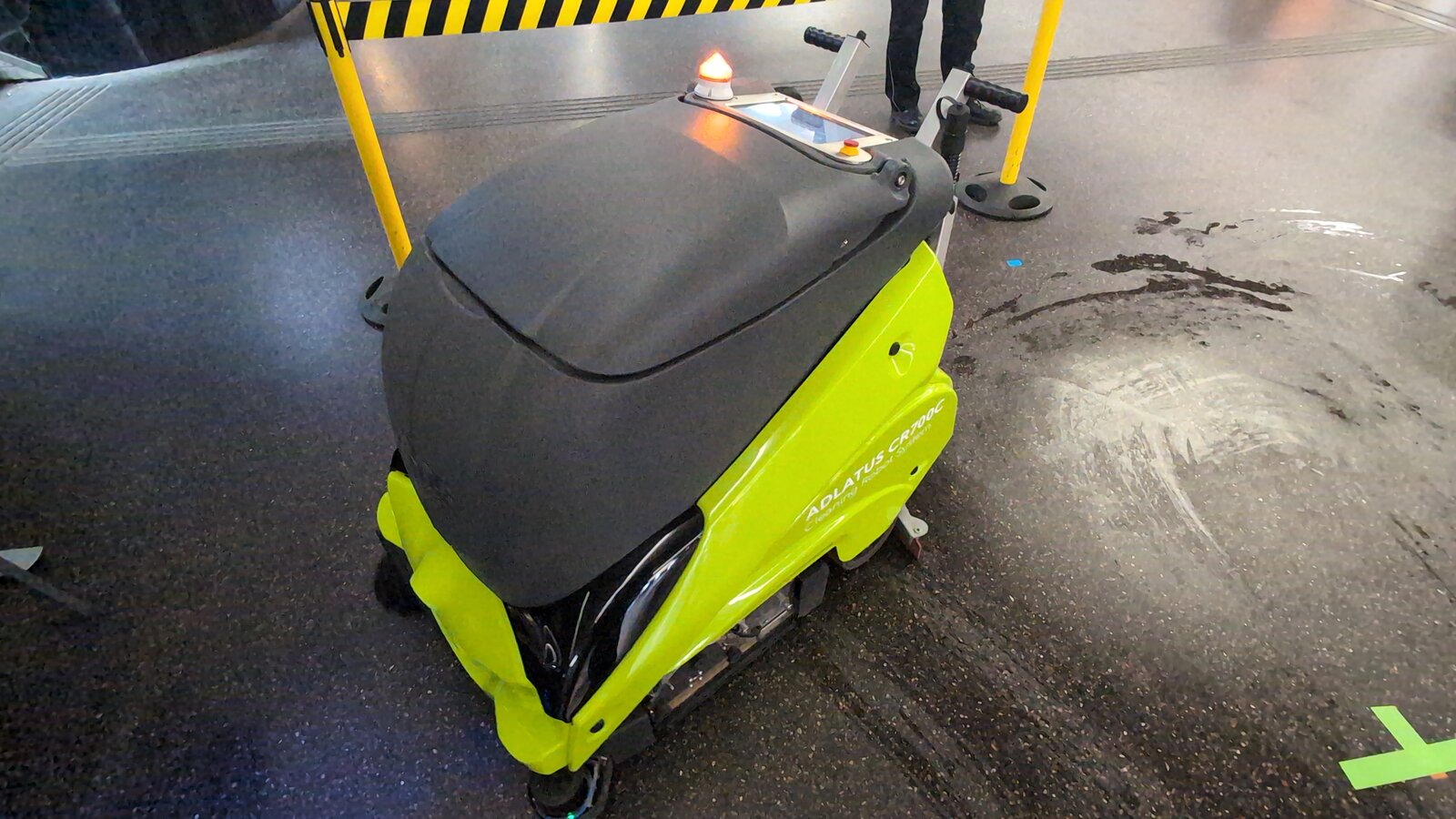}
    \hfill
    \includegraphics[height=.22\linewidth]{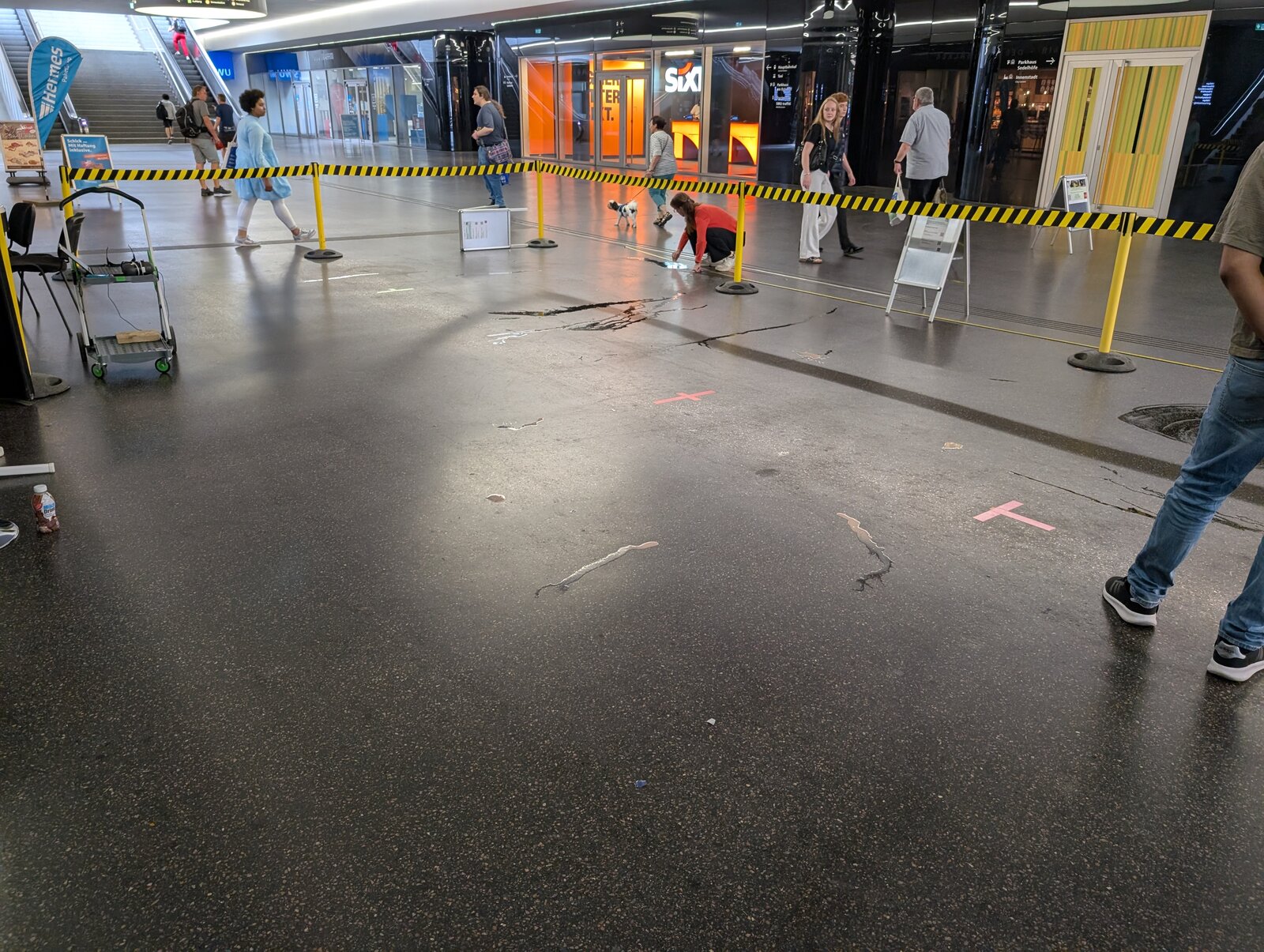}
    \hfill
    \includegraphics[height=.22\linewidth]{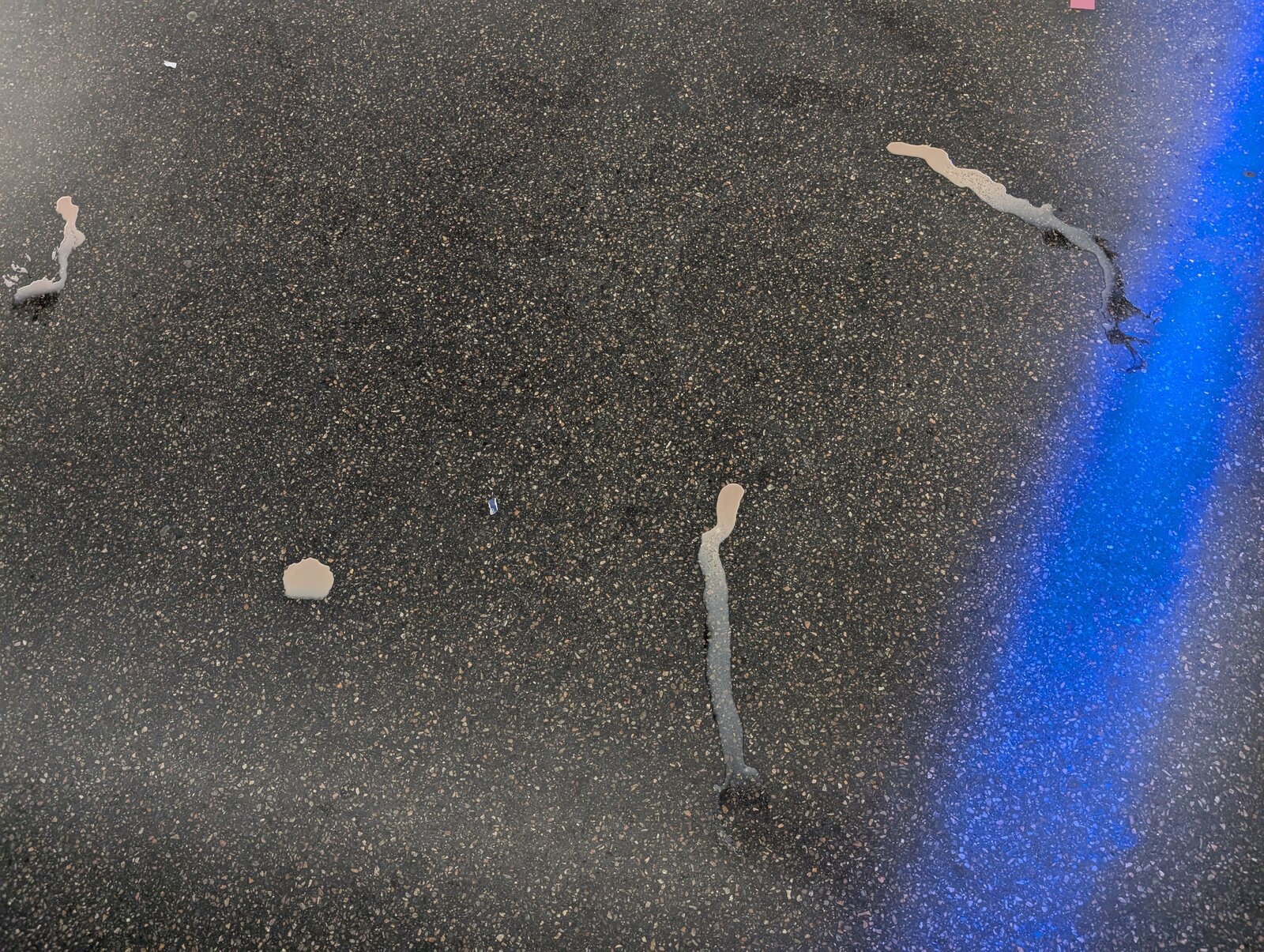}    
    \caption{Exemplary impressions from the Phase 2 evaluation of the \emph{Public Underpass Cleaning} use case, cleaning a restricted area at the Ulm main station underpass. The deployed cleaning robot (left), an overview of the restricted area (center) and exemplary coffee spills (right).}
    \label{fig:phase3_zen_mri_area}
\end{figure}
The Phase 2 evaluation of the \emph{Pedestrian Underpass Cleaning} case took place in a restricted area of the Ulm main station underpass (see \cref{fig:phase_2_overview}). Dirt particles, as defined in the evaluation concept, were distributed in the restricted area of \SI{6.5}{\metre} by \SI{10.4}{\metre}, yielding a total area to be cleaned of \SI{67.6}{\metre\squared}. In total, 10 paper pieces and 10 coffee spills were randomly distributed in the area. For the Phase 2 evaluation, we omitted the distribution of broken glass pieces. Two different individuals and a  piece of luggage were placed within the area to test the obstacle avoidance abilities.  \cref{fig:phase3_zen_mri_area} shows the variations during the evaluation.  The cleaning duration of the restricted area was 5:00 minutes (one cleaning run). During that time, 10/10 coffee spills had been cleaned successfully, and 9/10 paper pieces were absorbed, resulting in a total \emph{Completeness and Quality} of \SI{95}{\percent}. The calculated cleaning \emph{Efficiency} is \SI{13.5}{\metre\squared\per\minute} which would be a projected efficiency of \SI{810}{\metre\squared\per\hour}.

\subsubsection{Results from Phase 3 (Task Fulfillment)}

\subparagraph*{Results from Public Library (Task Fulfillment, Phase 3)}

\begin{figure} 

    \centering

    \includegraphics[height=.32\linewidth]{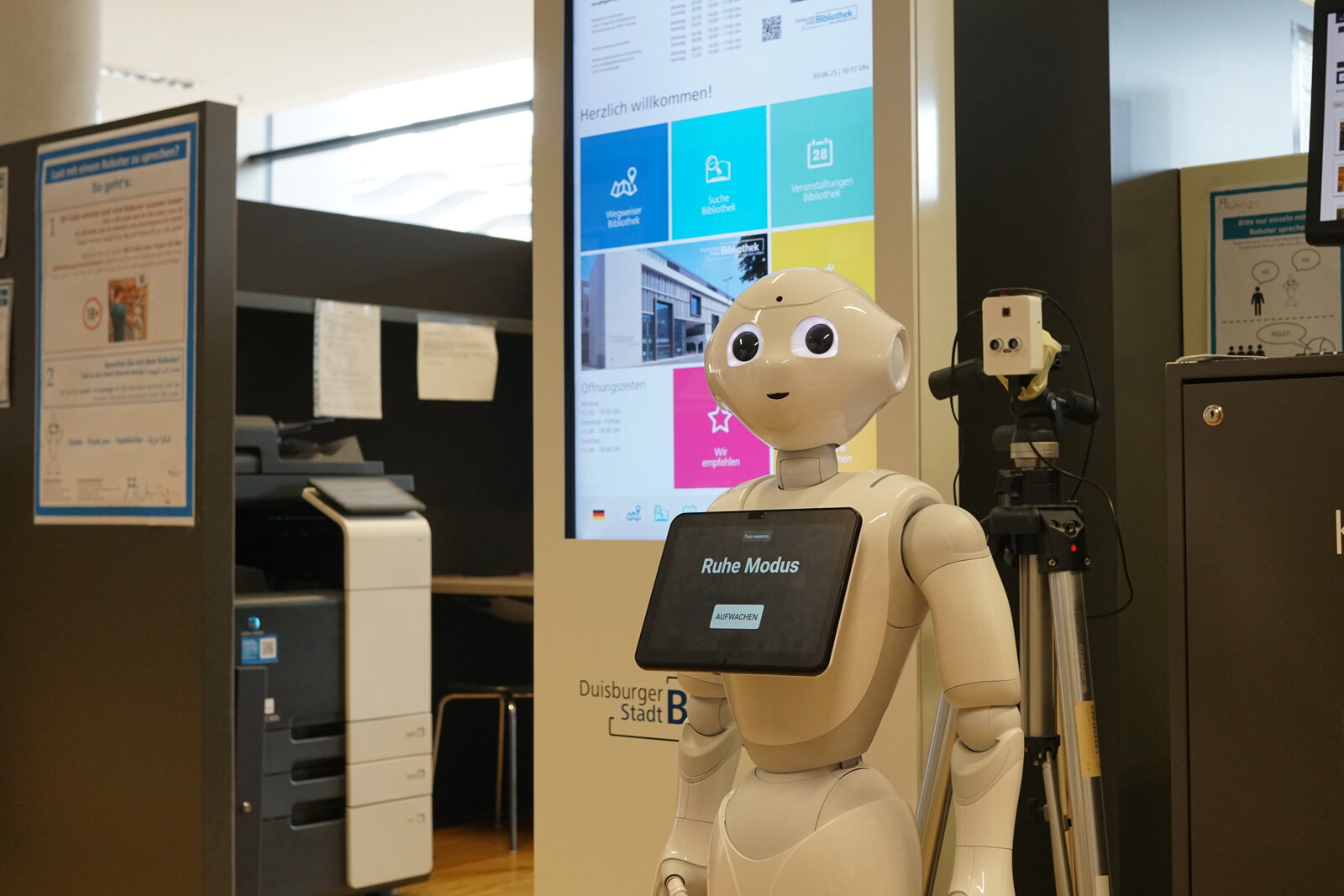}
    \hfill
    \includegraphics[height=.32\linewidth]{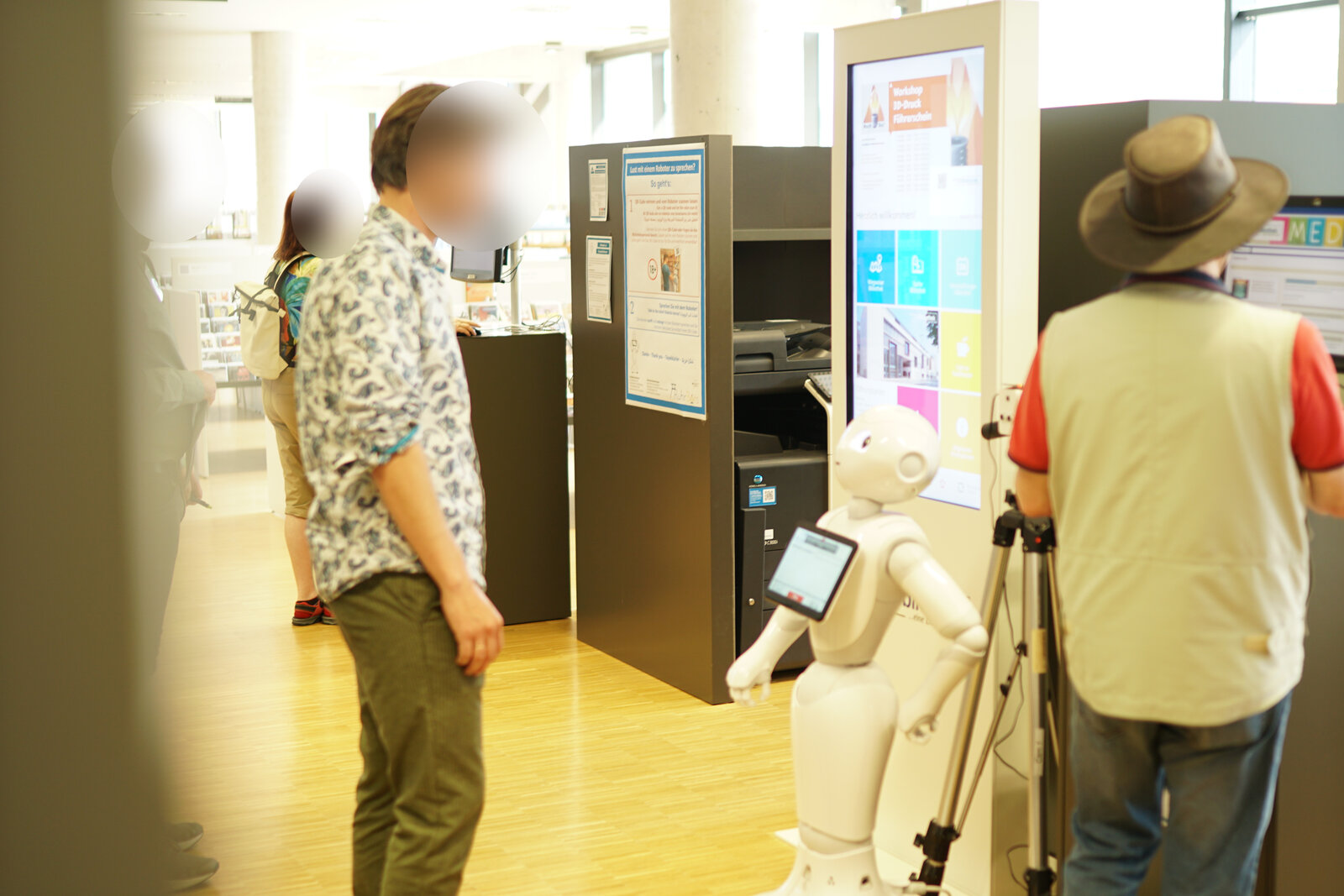}
    \caption{Exemplary impressions from the public library use case in the Duisburg public city library during the Phase 3 evaluation with the Pepper robot waiting for a user to approach (left) and within an interaction (right).}
    \label{fig:example_ruhrbots_phase3}
\end{figure}

For Phase 3, the \emph{Public Library} use case was assessed at the city library in Duisburg (see \cref{fig:example_ruhrbots_phase3}).
The evaluation followed the same protocol and utilized the same robots as outlined in \cref{sec:results-taskfulfil-library_phase2}. Passersby were invited to participate in the experiment as they showed interest; information boards installed in the library indicated the ongoing experiments (see Ethics Statement). One notable interaction involved a library visitor searching for a specific book. The dialog was initiated using a library ID card.
During the interaction, participants engaged with the robot naturally, which meant not all features were activated. While the visitor was welcomed, the interaction encountered several disruptions: the robot frequently misinterpreted commands and became stuck in prolonged processing routines. Although the visitor received the requested book information, the robot failed to provide the book’s specific location within the library. Additionally, the farewell sequence was not triggered.
Overall, the dialog success rate saw a modest improvement, reaching \SI{43.18}{\percent} (again based on a single systematically scored interaction, the one described above) compared to the Phase 2 experiment.


\subparagraph*{Results from Park Cleaning (Task Fulfillment, Phase 3)}

\begin{figure*}
    \centering
    \includegraphics[height=.25\linewidth]{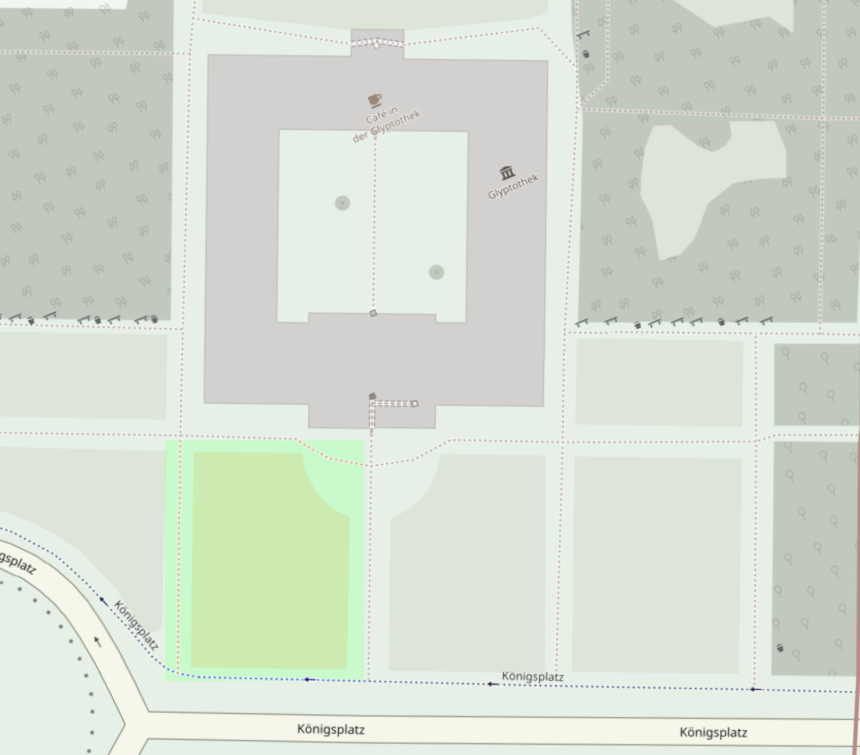}
    \hfill
    \includegraphics[height=.25\linewidth]{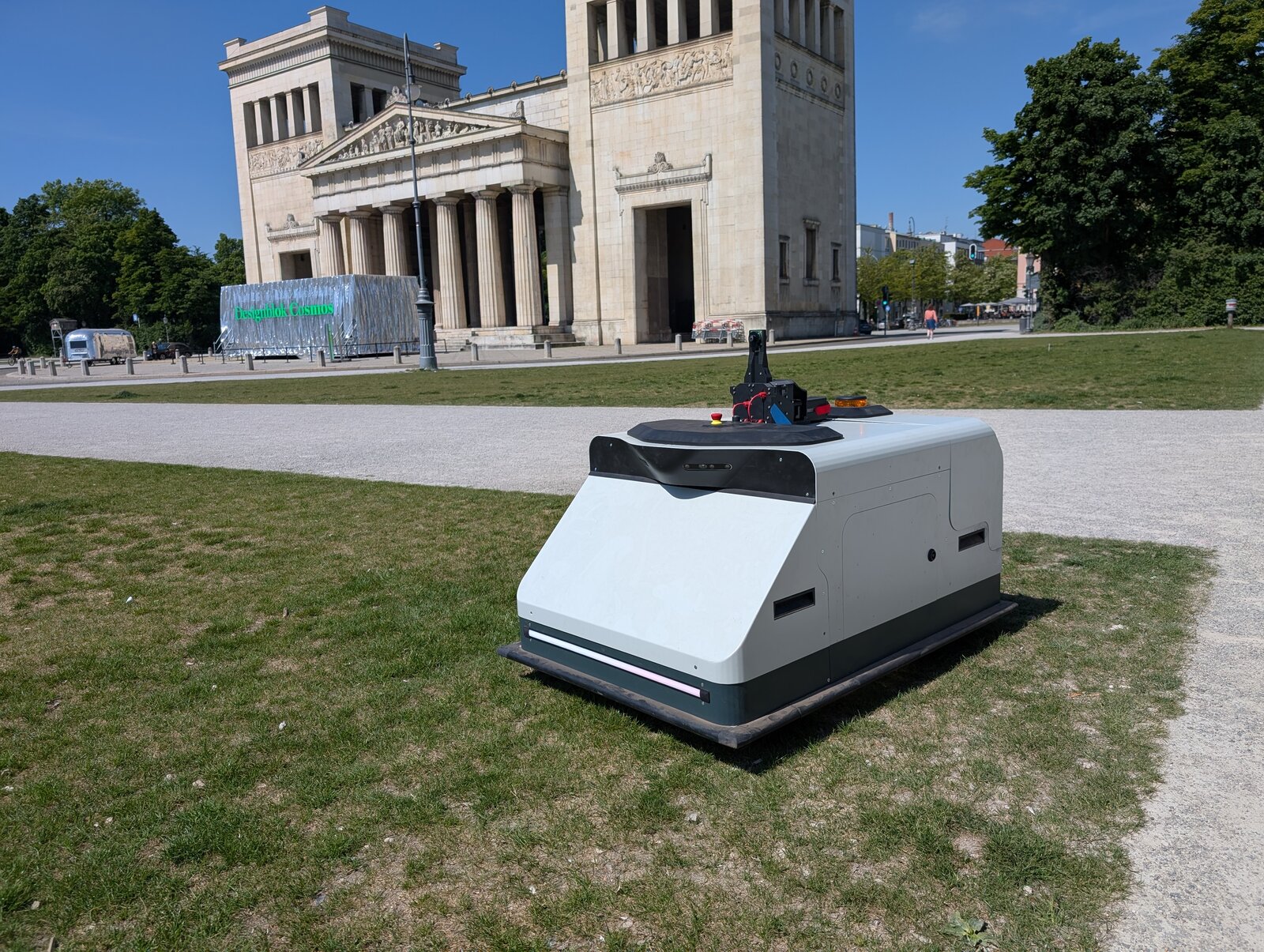}
    \hfill
    \includegraphics[height=.25\linewidth]{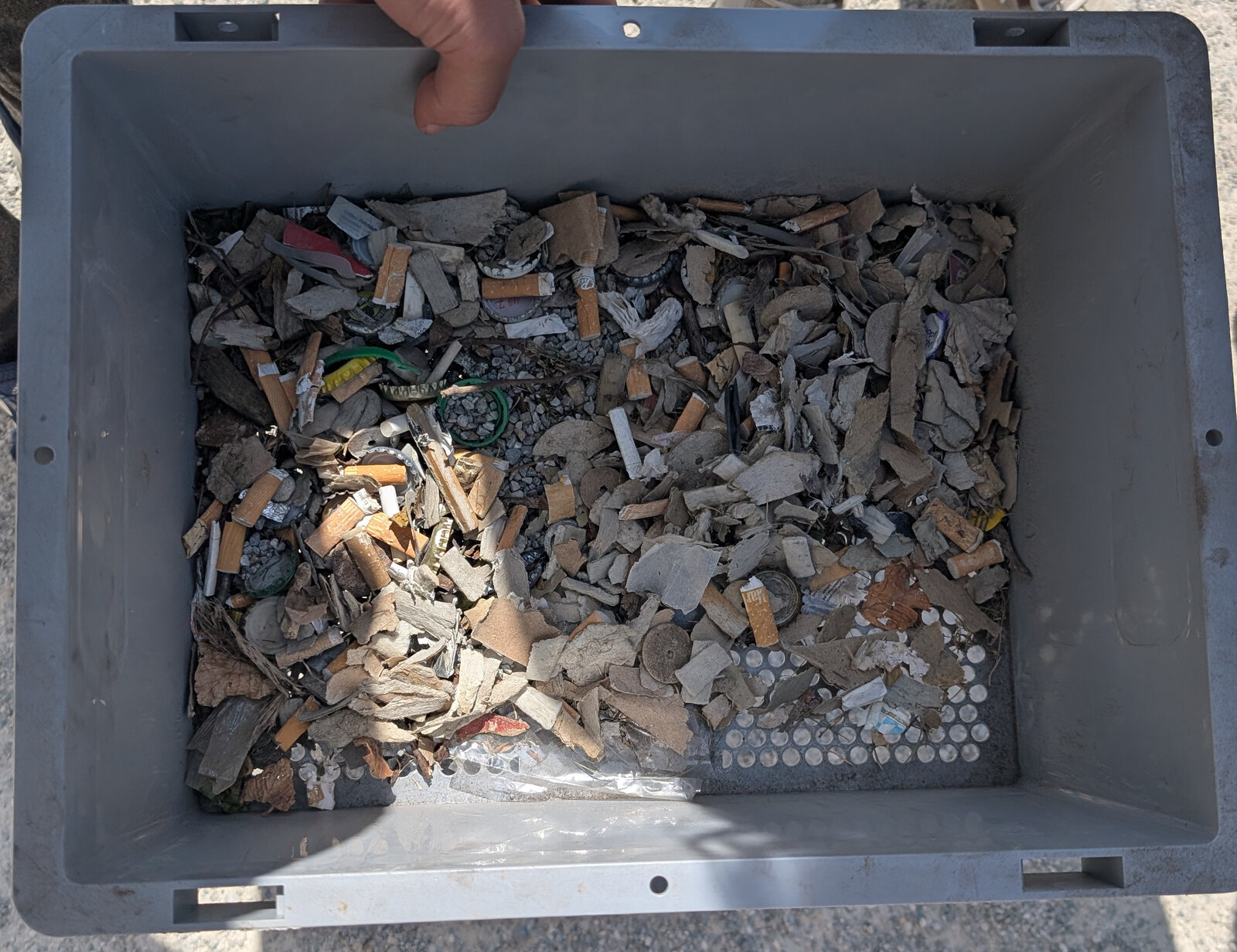}
    \caption{Task Fulfillment for the Park Cleaning Use Case. The area of the Glyptothek (Munich) that was selected for conducting the experiments (left) (© OpenStreetMap contributors), the robot cleaning the grass area during Experiment 1 (center), and the resulting task collection after Experiment 2 (right).}
    \label{fig:phase_3_rokit}
\end{figure*}

For Phase 3 of the park cleaning use case, the outdoor cleaning robot was publicly deployed in front of the Glyptothek museum in Munich. The highlighted area in \cref{fig:phase_3_rokit} (left) was selected for task fulfillment evaluation. Two experiments ($N=2$) were conducted during this phase.
In Experiment 1, the entire grass area was designated as the cleaning zone. The robot cleaned this area along the outer circle before proceeding to Experiment 2, which focused on a subarea for more systematic benchmarking. Trash items, primarily cigarette butts, were distributed across the designated area, and a hit-and-miss table was created, as in the Phase 2 evaluation. Impressions are depicted in \cref{fig:phase_3_rokit}.
For Experiment 1, most misses occurred because the robot avoided individuals, thereby bypassing areas where trash was located. Additionally, some bottle caps were embedded in the grass, preventing the vacuum mechanism from absorbing them.
In Experiment 2, no active disturbances were introduced, and the robot successfully collected all distributed trash. This experiment was conducted on a gravel surface, where bottle caps remained on the surface rather than being embedded in the ground.
Results of both experiments are shown in \cref{tab:phase3_rokit}.
Note that the internal project report counted 219 collected items for the same 41-minute run of Experiment 1 (row 1* in \cref{tab:phase3_rokit}), whereas our evaluation counted 23 items. This near tenfold difference reflects what was measured in each source: the panel's hit-and-miss evaluation considered only the panel-distributed ground-truth items within the designated evaluation area, whereas the robot's internal log counted every item collected across the entire, naturally littered public lawn --- predominantly cigarette butts already present in the park (206 of the 219 logged items). In addition, the internal report's automated classification also counted visually similar objects, such as rotten leaves, as cigarette butts, further inflating the reported number of cigarettes. The internal count therefore illustrates the realistic workload of the deployment, while only the panel-distributed items allow computing detection performance against a known ground truth.

\begin{table}

\centering
\begin{tabular}{c|c|c|c|c|c|c}
    \toprule
     \textbf{Experiment} & \textbf{Duration} & \textbf{Total} & \textbf{Cigarettes} & \textbf{Bottle Caps} & \textbf{Miss Cig} & \textbf{Miss Bottle Caps}\\
     \midrule
     1 & 41:00 & 23 & 10 & 6 & 3 &4\\
     2 & 5:31 & 15 &  10 &  5 & 0 & 0 \\ 
     1* & 41:00 & 219 & 206 & 13 & - &-\\
     \bottomrule
\end{tabular}
\caption{Phase 3 results from the \emph{Park Cleaning} use case in a public park in Munich. * Denotes results from the internal report, which counted all collected items including litter already present in the park; see the main text for details. }
\label{tab:phase3_rokit}

\end{table}

\subparagraph*{Results from Pedestrian Underpass Cleaning (Task Fulfillment, Phase 3)}

\begin{figure}

    \centering
    \includegraphics[width=.32\linewidth]{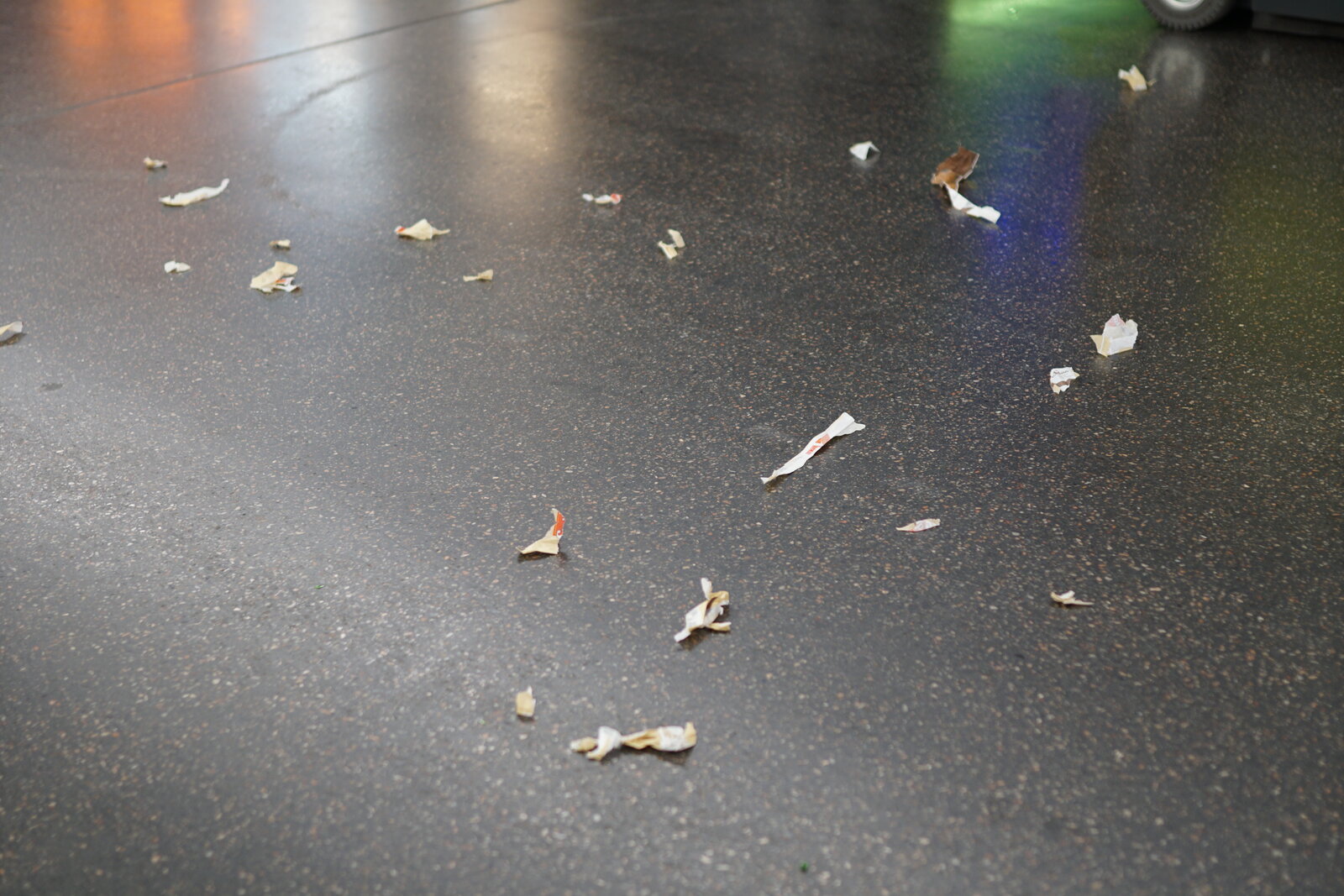}
    \hfill
    \includegraphics[width=.32\linewidth]{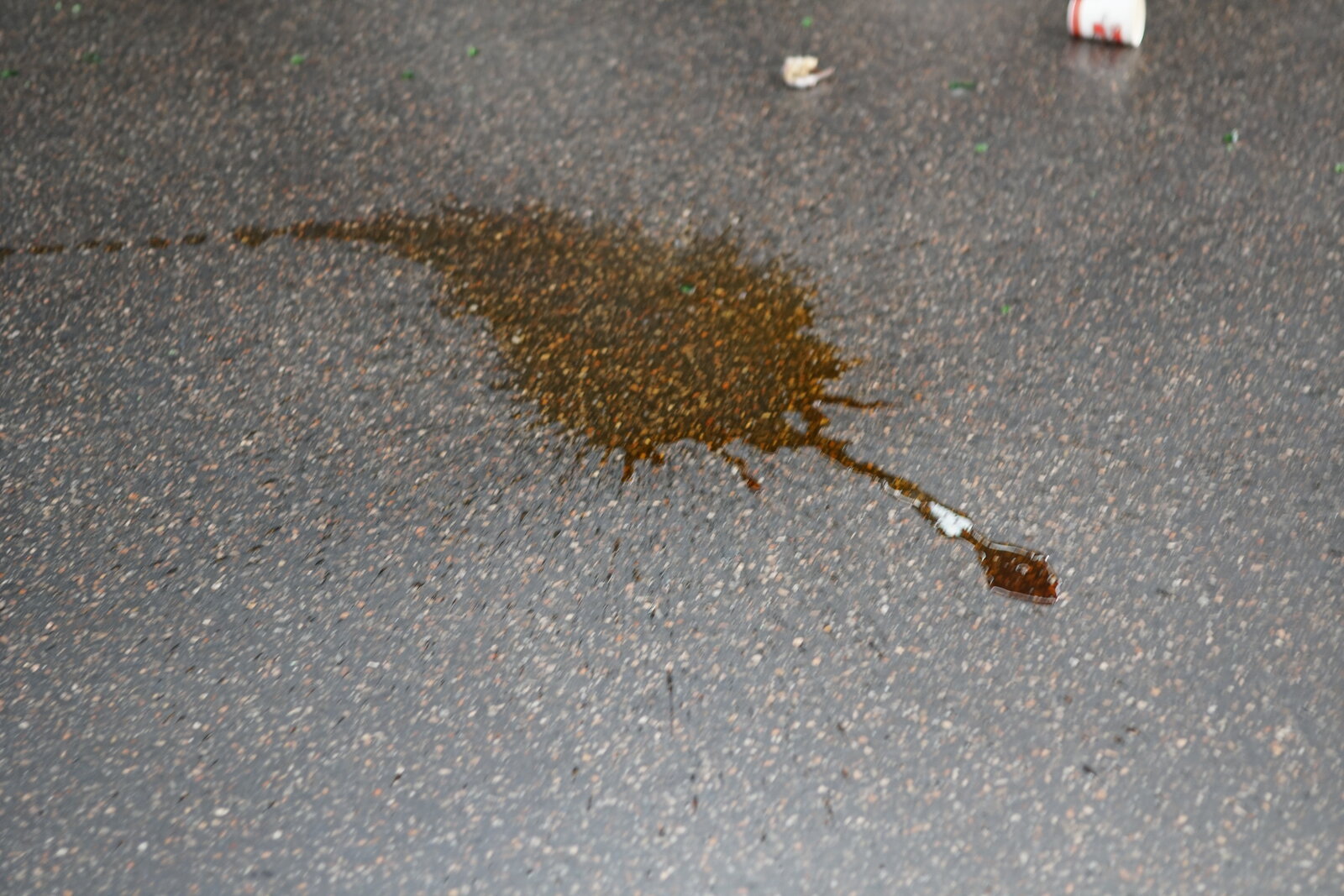}   
    \hfill
    \includegraphics[width=.32\linewidth]{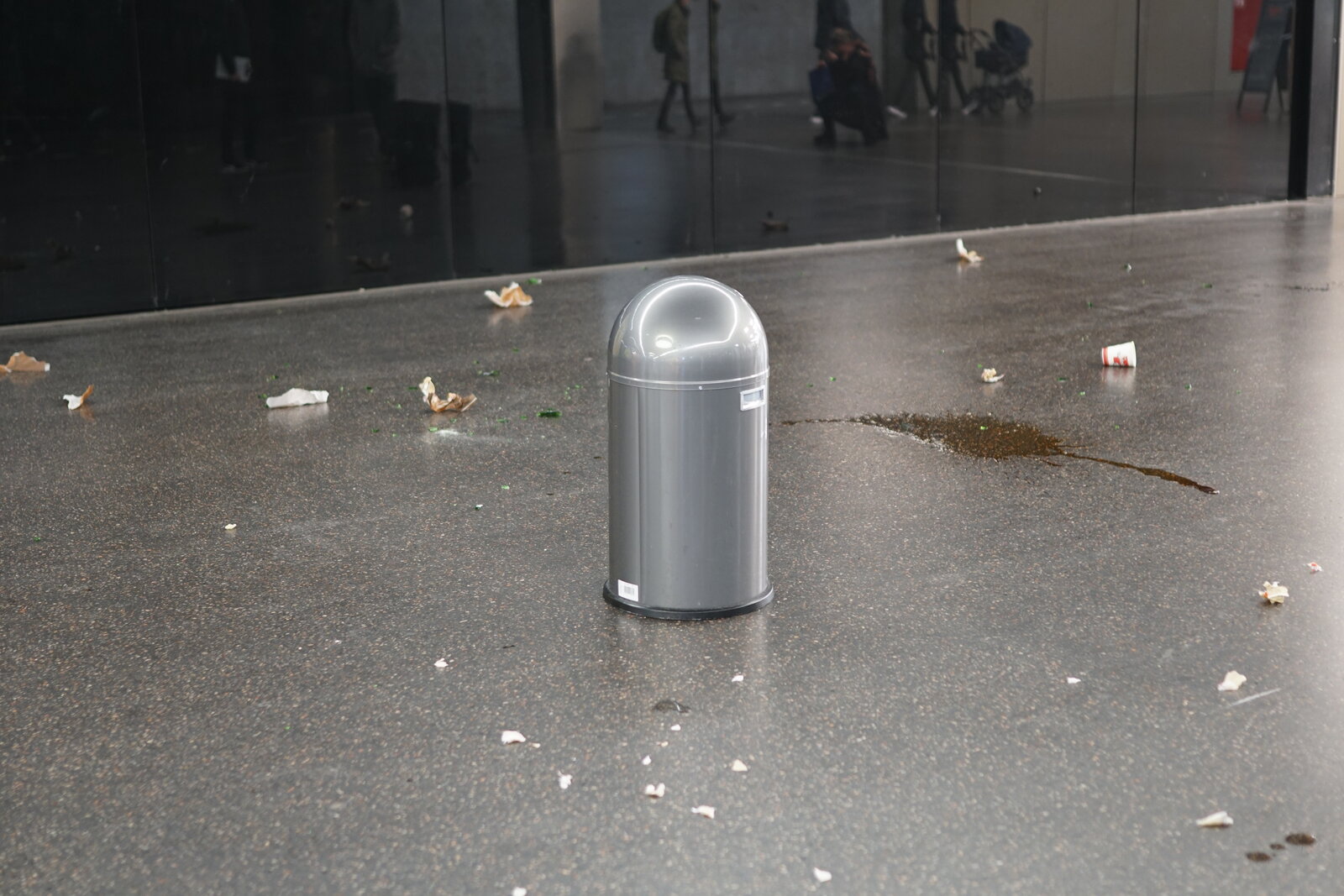}    
    \caption{Exemplary Impurities for the Pedestrian Underpass Cleaning Use Case.}
    \label{fig:example_impurities_zenmri_phase2}
\end{figure}

The pedestrian underpass cleaning use case was evaluated in an underpass at the main train station of Ulm. The cleaning robots are practically employed. For the task fulfillment, the \enquote{Completeness and Quality} as well as \enquote{Efficiency} were evaluated as proposed in the evaluation concept. In total, a measured area of \SI{7}{\metre} by \SI{18.5}{\metre} was defined for cleaning. The cleaning (one cleaning run) took 9~minutes, resulting in an efficiency of \SI{14.5}{\metre\squared\per\minute}. Obstacles in the form of humans, luggage, and bins were positioned. All installed obstacles were successfully avoided. Impurities in the form of coffee, paper pieces, broken glass, and paper cups were distributed in the defined area. The robot successfully handled the paper pieces, broken glass, and, up to a certain size, also the paper cups. The coffee stains were distributed instead of being cleaned.
Exemplary impurities of the task fulfillment evaluation are shown in \cref{fig:example_impurities_zenmri_phase2}.

\subsection{Interaction Quality Results}
This section presents the results of the benchmarking phases regarding interaction quality across the use cases. For each phase, the respective concepts, instruments, and observations are outlined, reflecting how interaction-related aspects were defined and examined under laboratory and real-world conditions.

\subsubsection{Results from Phase 1 (Interaction Quality)}
Within the benchmarking categories, interaction quality emerged as the dimension showing both the strongest conceptual commonalities and the most pronounced differences between the use cases. Phase 1 therefore focused on identifying shared principles while actively examining contextual divergences in the proposed metrics. Particular attention was paid to the testability, feasibility within the available time frame, and the facilitation of structured expert exchange during the consensus process.

\subparagraph*{Results from Public Library (Interaction Quality, Phase 1)}

In the library use case, interaction quality was addressed by metrics focusing on user experience during interaction rather than dialog success alone. Based on the consensus workshop, four metric dimensions were proposed: \emph{User Experience and Usability}, \emph{Consideration of Social Norms}, \emph{Need-Oriented Interaction}, and \emph{Diversity-Sensitive Interaction}.
\emph{User experience and usability} were defined based on general interaction principles applicable across interface types, as set out in ISO 9241-110:2020. In this use case, these principles mainly relate to social engagement, perceived usefulness, the quality of verbal communication, and the enjoyment of interaction. It was agreed that these aspects should be assessed using standardized questionnaires administered after task completion, complemented by analysis of recorded interaction data. The \emph{Consideration of Social Norms} addressed whether the robot behaved in a socially expected manner and was perceived as an appropriate interaction partner. A combined evaluation approach was proposed, consisting of qualitative interviews and questionnaire-based assessment of perceived social qualities, using established instruments where possible. \emph{Need-Oriented Interaction} was selected to capture user well-being during task execution, drawing on Self-Determination Theory. Perceived autonomy, competence, and relatedness were identified as key constructs and proposed to be measured using standardized post-interaction questionnaires. Finally, \emph{Diversity-Sensitive Interaction} was chosen to ensure that task quality remains robust across user groups. Diversity characteristics such as age, gender, and language proficiency were included as analytical factors to identify systematic differences and avoid unintended negative effects on interaction quality.

Overall, the consensus workshop emphasized the development and sharing of concrete questionnaire instruments, alignment across use cases where feasible.
 
\subparagraph*{Results from Park Cleaning (Interaction Quality, Phase 1)}

For the park-cleaning use case, metrics addressing \emph{Self-Descriptive Capability} and \emph{Existence Acceptance} (EA) were proposed and agreed upon in the consensus workshop to assess interaction quality during incidental encounters in public space. As passersby were assumed to have little to no prior knowledge of the robot system and no explicit usage intention, interaction quality was conceptualized in terms of immediate comprehensibility and acceptance of the robot’s presence.
The \emph{Self-Descriptive Capability} of the robot was defined as its ability to convey its role, task, and operational state through appearance, motion patterns, and acoustic cues. It was agreed that the robot should be presented to different user groups in varying environments and operational modes (e.g., moving, stationary, active cleaning). A questionnaire-based instrument was proposed to quantify the correspondence between the perceived and actual functionality of the robot, thereby capturing how accurately observers form a mental model of the system.
To complement this, \emph{EA} was identified as a suitable metric for non-instrumental interactions in public settings. In contrast to classical technology acceptance models, \emph{EA} integrates socio-emotional, cognitive, and interaction-related correlates of attitudes and behavior toward the mere presence of a robot. The benchmarking panel agreed that, depending on the study context, \emph{EA} may be measured using either a single-item indicator or a more comprehensive questionnaire reflecting its multidimensional structure.
Overall, the consensus workshop emphasized using standardized and, where possible, uniform questionnaires, testing with diverse participant groups, and evaluation under highly naturalistic conditions.

\subparagraph*{Results from Pedestrian Underpass Cleaning (Interaction Quality, Phase 1)}

In the third use-case, metrics were defined to evaluate the interactions between the robot system and passersby in the context of an underpass application. The focus was on \emph{trustworthiness}, \emph{consideration of needs for vulnerable groups}, \emph{efficiency of the trajectories for minimizing disruption}, and \emph{well-being of passersby}.
To measure \emph{trustworthiness}, the use of a trust scale was presented, which reflects both upstream trust beliefs (e.g., comprehensibility, reliability, predictability) and an integrative trust assessment. Online surveys and interviews were rejected as core instruments during the consensus workshop. The evaluation is quantitative, supplemented by structured observations of communication and interaction behavior. For \emph{consideration of needs for vulnerable groups}, it was decided to conduct tests with diverse user groups, using comparable trust values between groups as the metric. Video footage is used to analyze the \emph{efficiency of the trajectories for minimizing disruption}. The time needed to adjust behavior during conflicts, the duration until conflict resolution, and the crossing time of defined route sections relative to a baseline are evaluated.
The benchmarking panel emphasized the importance of using established, standardized questionnaires; ensuring a diverse, randomized selection of test subjects; and testing the instruments with test subjects in a real-world application context as planned.

\subsubsection{Results from Phase 2 (Interaction Quality)}
Building on the metrics defined in Phase 1, the evaluation concepts were further refined and explored in Phase 2 through practice-oriented testing. The three projects jointly developed an initial guideline for assessing interaction quality, intended as a structured checklist to support expert-based review of interaction concepts across use cases. The guideline aimed to identify a lowest common denominator while reducing the reliance on extensive user studies.
The guideline comprised categories derived from established usability and interaction principles, including \emph{visibility of system status}; \emph{match between system and the real world}; \emph{controllability, user control and freedom}; \emph{consistency and standards}; \emph{error prevention}; \emph{recognition rather than recall}; \emph{flexibility and efficiency of use}; \emph{aesthetic and minimalist design}; \emph{support for error recognition and recovery}; \emph{help and documentation}; \emph{ISO 9241 principles}; \emph{robot trustworthiness}; and \emph{accessibility}. The document represented an instrument informed by prior literature and practical experience and was explicitly understood as such.

\subparagraph*{Results from Public Library (Interaction Quality, Phase 2)}

For the public library use case, Phase 2 evaluation was conducted in a laboratory environment. Unlike the other scenarios, a use-case-specific retrieval for the guideline was employed that was tailored to the categories of \emph{user experience and usability}, \emph{consideration of social norms}, \emph{need-oriented interaction}, and \emph{diversity-sensitive interaction} of the evaluation concept.
The benchmarking panel interacted directly with the robot and observed the interactions of other members. Recurrent issues included long processing times, interaction breakdowns following misunderstandings by the large language model, and dissatisfaction with the resulting dialog flow. Additionally, non-adaptive gestural behavior was identified as a source of distraction, and inconsistent gaze behavior, which was sometimes directed toward bystanders rather than the interaction partner, was found to negatively impact the perceived interaction quality.

\subparagraph*{Results from Park Cleaning (Interaction Quality, Phase 2)}

For the park cleaning use case, the shared guideline was used as the basis for a questionnaire with particular emphasis on \emph{self-descriptive capability} and \emph{inclusivity and accessibility}. Following the interim consensus meeting after Phase 2, the latter replaced the \emph{EA} category.
Each member of the benchmarking panel received the questionnaire and interacted with the robot under controlled conditions. While the guideline provided a comprehensive set, several aspects were found to be difficult to operationalize for this use case. Observations revealed, among other aspects, that the robot occasionally performed rapid and pronounced turning maneuvers, requiring persons to quickly leave its turning radius. In addition, status indications via color coding were discussed as not always being clearly understandable. It was further noted that some guideline categories and formulations were not readily understandable or assessable for non-expert users, limiting their applicability for broader evaluation.

\subparagraph*{Results from Pedestrian Underpass Cleaning (Interaction Quality, Phase 2)}

For the pedestrian underpass cleaning use case, the guideline was used focusing on the categories \emph{Usability}, \emph{Robot Trustworthiness}, and \emph{Accessibility}. 
The evaluation involved selectively invited participants to examine when interactions became uncomfortable. Testing was conducted within a cordoned-off area, while the benchmarking panel observed participant behavior from a distance. Similar to the park cleaning scenario, it was found that several guideline-based categories were not applicable or not directly assessable in this spatially constrained and highly dynamic setting.

\subsubsection{Results from Phase 3 (Interaction Quality)}
In Phase 3, the interaction quality categories in the evaluation concepts developed and refined in the previous phases were applied without further modification. Nevertheless, changes in the questionnaires were allowed. The focus of this phase lay on examining how the defined interaction metrics performed under real operational conditions in each use case. 

\subparagraph*{Results from Public Library (Interaction Quality, Phase 3)}

In the public library use case, the predefined categories were applied in the library during operating hours.
Real-world environmental factors affected the quality of interactions. Speech recognition performance was affected by network instability and ambient noise, which occasionally led to misunderstandings and reduced dialogue coherence. Additionally, the low visitor traffic during testing seemed to increase users' self-awareness during interaction. The lack of integration with the real library catalog limited the perceived usefulness of interacting with the robot.
The benchmarking panel again supported structured evaluation. Spontaneous users were not asked to complete the questionnaire.

\subparagraph*{Results from Park Cleaning (Interaction Quality, Phase 3)}

For the park cleaning use case, the revised interaction questionnaire developed after Phase 2 was applied during public deployment. The instrument comprised six overarching categories — \emph{transparent}, \emph{purposeful}, \emph{robust}, \emph{defensive}, \emph{conventional}, and \emph{accessible} — each operationalized through sub-questions. In addition to qualitative comment fields, evaluators rated each category on a 10-point scale ranging from “not fulfilled” to “fulfilled”.
During deployment in the public setting, the robot was clearly perceived by passersby. Individuals occasionally stopped to observe the system or took photographs; however, close or sustained interactions between pedestrians and the robot occurred rarely. Most passersby maintained distance and continued their activities without direct engagement.
The structured questionnaire was completed exclusively by members of the benchmarking panel, based on observation and situational interaction tests conducted during deployment. At no point were passersby asked to complete or apply the questionnaire themselves.

\subparagraph*{Results from Pedestrian Underpass Cleaning (Interaction Quality, Phase 3)}

In the pedestrian underpass use case, the established categories were applied in the operational environment.
Although pedestrian traffic continued during the tests, many people avoided getting too close, partly due to the marked area around the robot's workspace. Members of the benchmarking committee interacted with the robot to test its boundary conditions and robustness.
Panel feedback indicated that combining multiple signaling modalities (color elements, blinking lights, projected indicators, icons, and speech output) could lead to perceptual overload in the confined space of the underpass. Yet multimodal HRI communication is indicated to design for accessibility and for distracted pedestrians; a balance between potential overload and accessibility is therefore needed. The comprehensive questionnaire allowed for detailed assessment but required substantial time and simultaneous consideration of multiple criteria. As in the other scenarios, passersby did not apply the instrument themselves.
\subsection{Safety Results}
\subsubsection{Results from Phase 1 concerning Safety}
The safety benchmarking started with written suggestions of what to look for safety-wise by the leading personnel of each use case within a scope that was set by safety experts from the benchmarking panel (see \cref{subsec:phases}). The suggestions were then discussed with everyone involved during a consensus workshop in Bonn with the goal of finding common ground with all projects, use cases, robot applicators, benchmarking panel for the first test (see \cref{tab:phase_exec}). Discussions revolved largely around the testability in the field. Everyone, e.g., agreed to test the existence and function of an emergency stop button on each robot. As noted in \cref{sec:relatedwork-safety}, two main issues had to be focused on: First, there are few, if any, applicable standards for robots in public spaces. Second, public spaces pose problems beyond the scope of ordinary industrial standards that may demand stricter safety measures. 

\subparagraph*{Results from Public Library  Scenario (Safety, Phase 1)}

For the robot in the public library use case, the testing of electrical safety was discussed but dismissed as not feasible during a specific test run during a benchmarking event. The discussions again focused instead on functional and mechanical safety. As the robot was, in the end, not expected to be moving, the risk of colliding with a human, impact forces, and safely stopping in case of an incident was considered minimal. Still, the safety stop was discussed as a relevant safety function. ISO 13482:2014~\citep{ISO13482} and ISO/TS 15066:2016~\citep{ISO15006} were agreed on to be used as core safety standards. 

\subparagraph*{Results from Park Cleaning Scenario (Safety, Phase 1)}

For the park cleaning robot, the testing of electrical safety was discussed but dismissed as not feasible during a specific test run during a benchmarking event. The discussions focused instead on mechanical safety -- like the risk of colliding with a human, impact forces, and safely stopping in case of an incident. On the sensory side, the discussed robot relies on an RGB-D camera and a bumper for detecting individuals. Only the bumper is involved in a safety function. Yet, the benchmarking panel also investigated the RGB-D camera because it allows the demonstration of interesting challenges for robots in public spaces.

To investigate the detection capability of an imagined safety function that involves the RGB-D camera, the collision avoidance behaviour was triggered via the camera system. Test pieces in lieu of real individuals were agreed on to avoid endangering individuals during these tests. For the test pieces, the following requirements were decided on: they should be representative of the operational design domain to produce meaningful results, be standardized to foster reproducibility and fairness of examinations, and be easy to use and quick to set up (for example, shape symmetry allows for easy positioning).
\begin{table}[tbp]

\centering
\begin{tabular}{@{}llp{9cm}@{}}
\toprule
\textbf{Test Piece}      & \textbf{Surface Material} & \textbf{Shape} \\
\midrule
Standing person          & Molleton or denim         & Upper part: cylinder (height 53\,cm, diameter 12\,cm); Lower part: frustum (height 47\,cm, diameter bottom 4.5\,cm, diameter top 12\,cm) \\
Squatted down person     & Molleton or denim         & Cylinder (height 50\,cm, diameter 30\,cm) \\
Hiking pole              & Molleton or denim         & Pipe (length 125\,cm, outer diameter 2\,cm) \\
Leg of charcoal grill & Chrome                    & Pipe (length 55\,cm, outer diameter 1.1\,cm) \\
\bottomrule
\end{tabular}
\caption{Main test pieces for safety benchmarking in park use case.}
\label{tab:results-safety-testpieces}
\end{table}
With the discussed issues regarding applicable standards for public spaces in mind, the following test pieces were created (see \cref{tab:results-safety-testpieces}): a standing and a squatted-down five-year old child; a hiking pole that, if ignored by the robot, may lead to a person falling; a leg of a charcoal grill represented by a chrome pipe that is difficult to detect, even by safety laser scanners, due to its thinness and reflective surface material. With the exception of the charcoal grill leg, all these test pieces are covered either in molleton, a low-reflective cloth, or denim that is more reflective, yet frequently encountered in public spaces.

\subparagraph*{Results from Pedestrian Underpass Cleaning  Scenario (Safety, Phase 1)}

For the robot in the pedestrian underpass use case, the testing of electrical safety was discussed but dismissed as not feasible, the same as in the case of park cleaning. The discussions again focused instead on functional and mechanical safety. The discussions varied from the park cleaning scenario, as the robot was equipped with a laser scanner instead of a bumper and a camera, among other reasons, due to their size and weight difference. The benchmarking panel agreed to benchmark the object detection through a laser scanner, the emergency stop through the e-stop button, and a drop test on a stair edge. DIN EN IEC 63327 VDE 0700-327:2023-03~\citep{IEC63327} was used as a core safety standard.

\subsubsection{Results from Phase 2 concerning Safety}
The evaluation concepts were then revised for each use case regarding the first test within a controlled environment. These tests were quite sufficient already regarding safety.

\subparagraph*{Results from Public Library Scenario (Safety, Phase 2)}

The library use case (Pepper robot) was tested in a laboratory environment in Bottrop. As Pepper comes ready with CE marking and in the end didn’t actually navigate through the laboratory (the robot was static and attached to a charging device at all times), no complex test cases were possible or necessary. 

\subparagraph*{Results from Park Cleaning Scenario (Safety, Phase 2)}

\begin{figure*}
    \includegraphics[width=.32\linewidth]{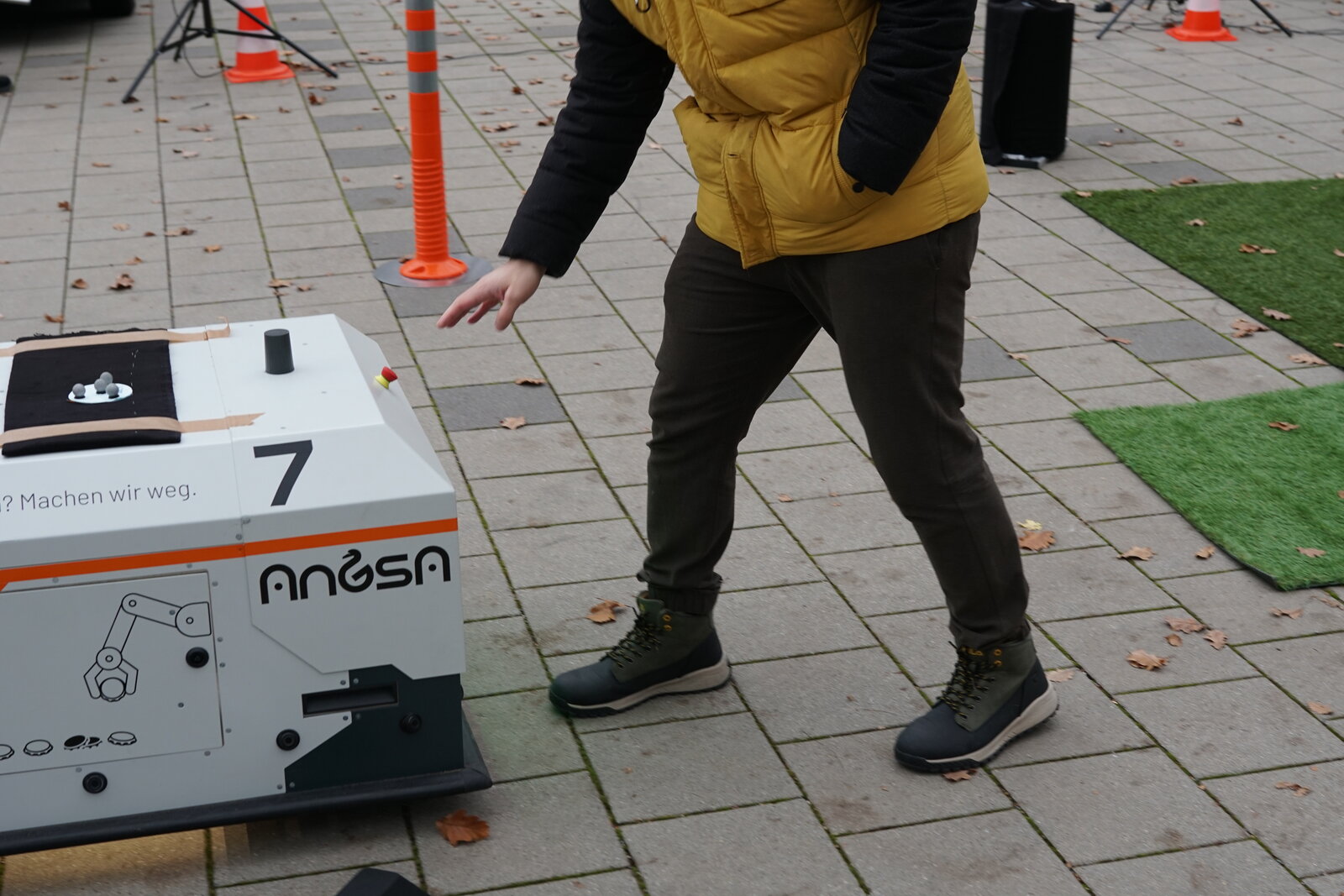}
    \hfill
    \includegraphics[width=.32\linewidth]{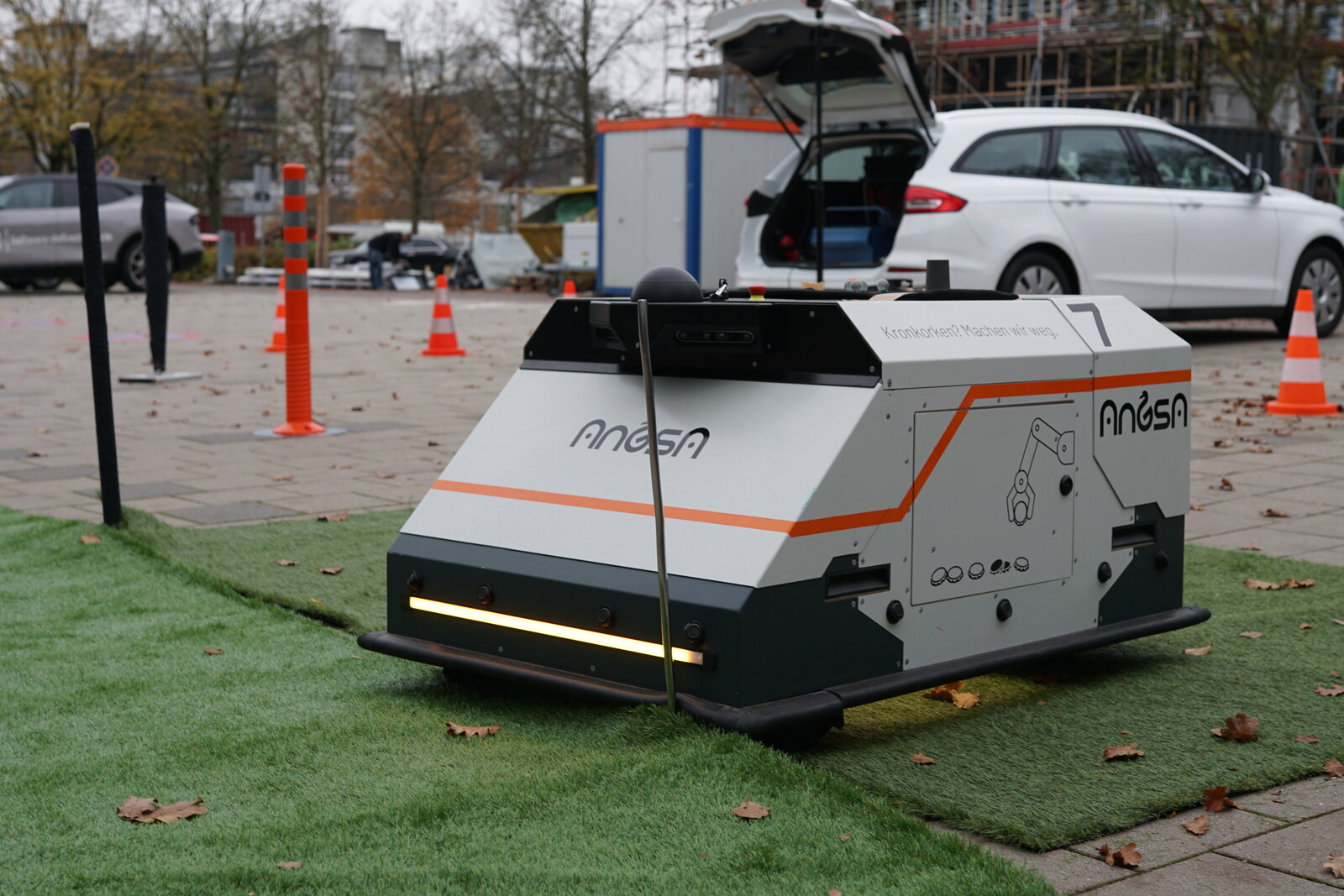}
    \hfill
    \includegraphics[width=.32\linewidth]{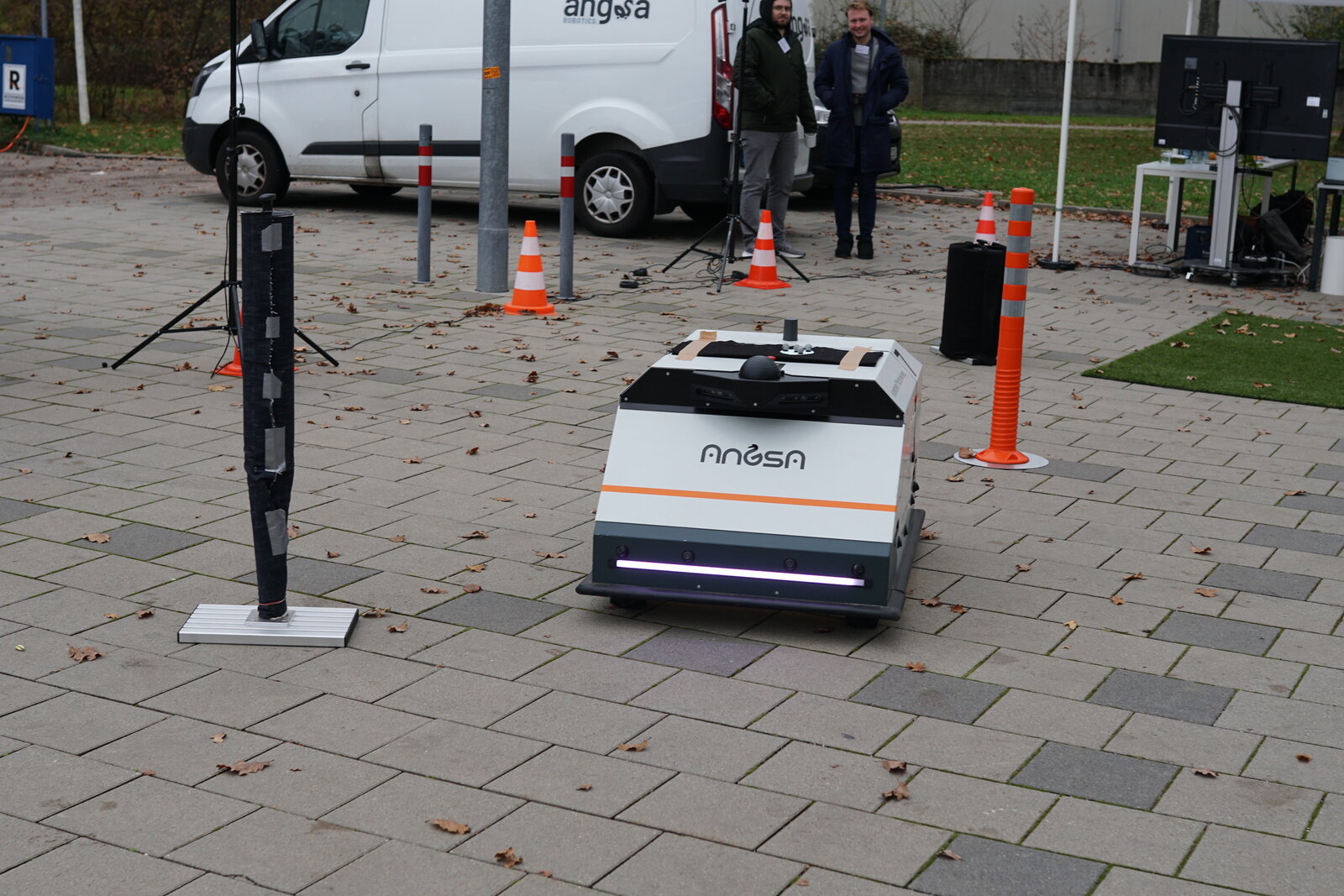}
    \caption{Safety test setup for controlled park cleaning scenario: Safety stop reaction (left), failed visual detection of the charcoal grill leg (center), detection of child standing up (right).}
    \label{fig:rokit-detection}
\end{figure*}

During the second phase, the robot's mechanisms were tested as planned in Phase 1 in a controlled test area with a fake lawn in Arena 2036 in Stuttgart. The test devices were used to benchmark the camera and the bumper, while only the bumper is involved in a safety function. The camera was tested regarding the detection of a standing and a squatted-down five-year-old child as well as a hiking pole and the leg of a charcoal grill (see \cref{tab:results-safety-testpieces}); the corresponding test setups are shown in \cref{fig:rokit-detection}. While the robot successfully detected the objects in most instances, the lack of a safety-rated camera restricts the collision avoidance behavior to a non-certified, purely functional operation.

\begin{figure*}
    \includegraphics[width=.32\linewidth]{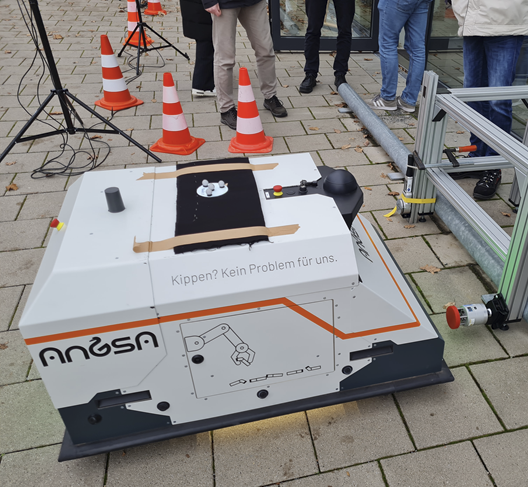}
    \includegraphics[width=.32\linewidth]{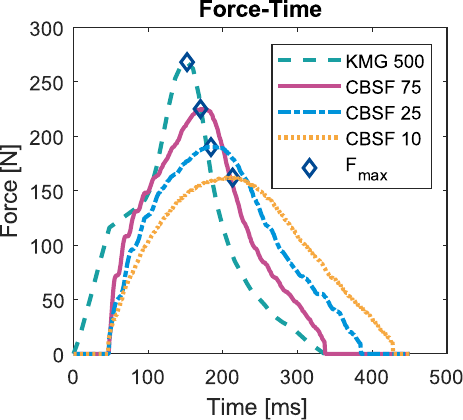}
    \includegraphics[width=.32\linewidth]{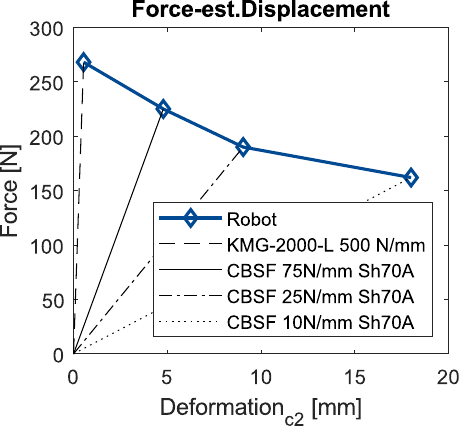}
    \caption{Collision test setup (left), Force-Time curves of single measurements (center), combined Force-est. Displacement Behaviour (right) with a target velocity of \SI{0.3}{\metre\per\second}.}
    \label{fig:rokit-collision}
\end{figure*}

This underscores the criticality of the bumper system as one of the robot's safety-rated protective devices and motivated a more comprehensive assessment. As part of the metric for analyzing mechanical hazards, possible collisions, such as with a leg, were examined. To this end, the robot was deliberately driven into biofidelic measuring devices at a target speed of 0.3 m/s (see \cref{fig:rokit-collision}). Such devices are also employed in the safety evaluation of industrial collaborative robots, automatic doors, or gates \citep{zimmermann2025}. More specifically, four measuring devices (GTE Industrieelektronik GmbH, Germany) – a KMG-2000-L and several CBSF-Devices (CoboSafe) – were used for this purpose. The test setup is shown in \cref{fig:rokit-collision} on the left. Examples of the measured data force over time, for the four devices, are shown in the middle. On the right, the result is shown as the curve of the maximum force over the estimated displacements for the different stiffnesses of the test devices; this could show the stopping distance of the robot in human tissue, depending on the body location's elasticity. The benchmarking panel also noted that while the autonomous mode of the robot targeted a speed of 0.3 m/s, the teleoperated mode was faster and exceeded the acceptable impact force regarding safety.

The benchmarking panel also tested the emergency stop through activation of the e-stop button. The robot also passed this test: after activating the emergency stop, the system stopped immediately.

\subparagraph*{Results from Pedestrian Underpass (Safety, Phase 2)}

\begin{figure*}
    \centering
    \includegraphics[width=.6\linewidth]{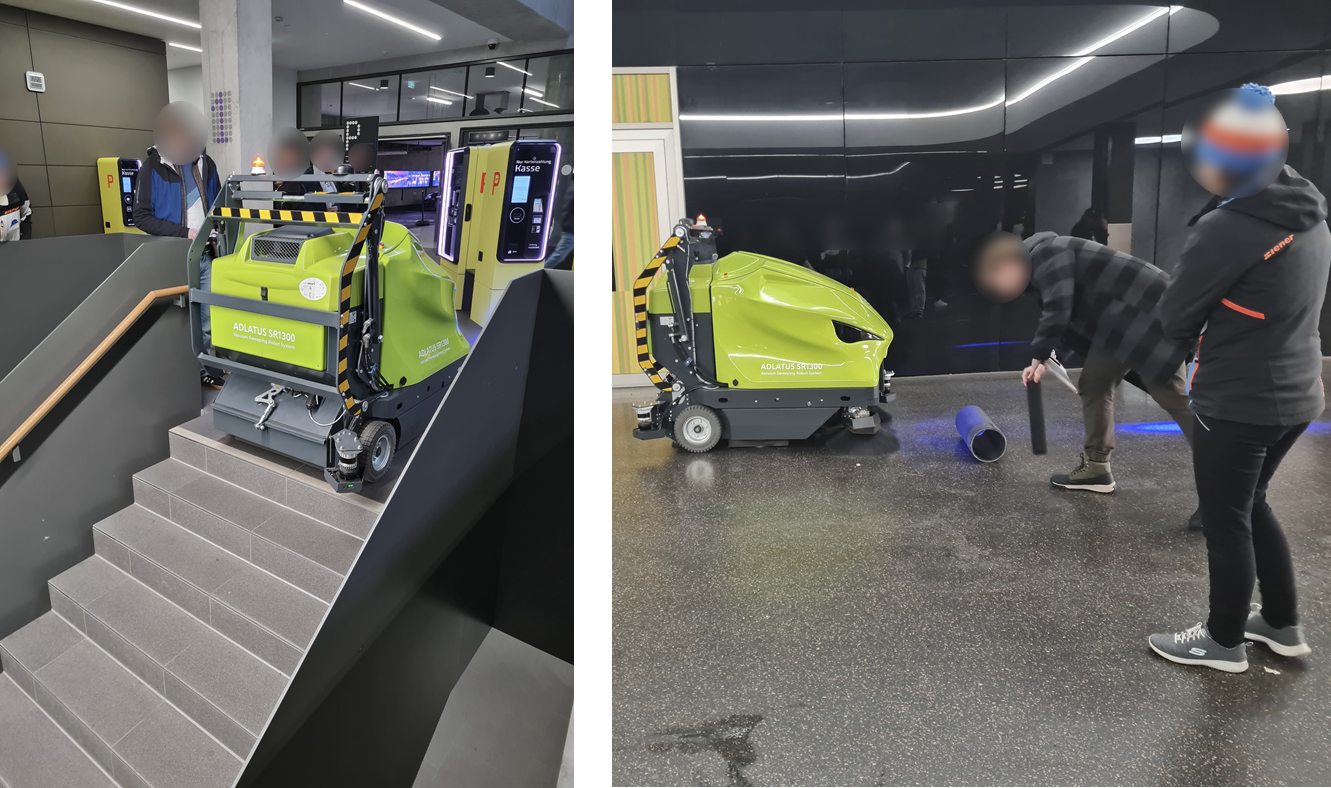}
    \caption{Safety test setup for controlled underpass cleaning scenario: drop test for stair detection (left) and detection of an adult with a test object (right).}
    \label{fig:safety_zenmri}
\end{figure*}

The underpass use case (Adlatus robot) was tested in a secured area in a subterranean area in Ulm (see \cref{fig:safety_zenmri}). A general calculation of functional safety was not done -- only functional tests were conducted. 

The benchmarking panel first tested the object detection through the attached certified laser scanner on the front side. The robot passed the test: It could detect different objects and stopped in time during a test in accordance with the black cylinder test for IEC 63327. Secondly, the benchmarking panel tested the emergency stop by activating the e-stop button. The robot also passed this test: After activating the button, the robot stopped immediately. Thirdly, we did a drop test on a stair edge. The robot passed, again. The robot stopped in front of the stairs and detected the edge with its laser scanner.

\subsubsection{Results from Phase 3 Scenario}
While all the other categories had tests in phase 3, the benchmarking panel agreed to be done with safety tests after phase 2, as no changes in the robots were made. 

\subsection{Results from Economic Viability}
\label{ssec:results_economic}
As introduced in \cref{sssec:category_economic}, the evaluation in this category followed phase-specific completion criteria. The economic benchmarking did not aim to produce comparative ratings of the three competence centers but to document how the toolbox-based approach unfolded in practice and what was learned about the methods themselves. Reported below are observations from the benchmarking reports, the three on-site Phase 2 sessions, the consensus workshop and interim meetings, and the three Phase 3 benchmarking events. A central observation across cases concerns the markedly different levels of prior expertise in business model development at the outset, which shaped the depth and trajectory of the economic work in each center and surfaced concrete needs for toolbox refinement.

\subsubsection{Cross-Case Observations}
Across the three competence centers, the benchmarking surfaced a recurring pattern: generic business model templates required contextual adaptation to capture the specifics of service robotics in public spaces. Two directions emerged consistently: domain-specific extensions where conventional methods did not adequately reflect human-robot interaction (e.g., the Human-Robo Journey as an extension of the Customer Journey), and value-logic adaptations where standard revenue assumptions did not apply, particularly in public-sector deployments (e.g., a Non-Profit Business Model Canvas variant). Insights of this kind were systematically fed back into the RimA Toolbox, leading to concrete refinements such as the integration of Target Pricing and Target Costing, the Non-Profit BMC adaptation, and the restructured collaborative tool guides developed after the  workshop on 17 June 2024. The toolbox thus operated as an iteratively refined research artifact, shaped through cyclic application and feedback. Across phases, the analytical scope shifted from the micro level of individual robotic applications toward meso-level considerations of center stabilization and, in one case, macro-level questions of platform-based scaling.

\subsubsection{Application-Oriented Operationalization of the Toolbox within the RoboSpot Ecosystem - a RuhrBots Example}
The RimA Toolbox is operationalized within the RoboSpot ecosystem, the economic core of the RuhrBots project, systematically translating scientific insights into a platform-based B2B business model for social robotics in public and business contexts. Its adaptive, iterative development integrates stakeholders across all phases through four key methodological components:
The Human–Robot Journey (HRJ) extends the Customer Journey framework, embedding social robots as autonomous actors within a three-layered model (persona profiles, experience environments, and service phases). Empirically validated through literature reviews and field studies (qualitative interviews, n = 65; surveys, n = 148), it supports structured evaluation and planning of deployment scenarios~\citep{RothKlicicBeder2025,RothKlicicBeder2026HRJ}.
Stakeholder Network Analysis refines the Toolbox’s stakeholder perspective into a formal governance framework, positioning public institutions and B2B actors within a platform architecture. Using 13 B2B personas, it differentiates supply/demand segments, defines value-creation roles, and assesses governance and revenue models, resulting in a coordinated platform with clear incentives~\citep{RothBederKlicic2026Brand}.
The AI-augmented Business Model Canvas structures scenario generation using organizational, transactional, and contextual parameters, incorporating Trustworthy AI principles~\citep{RothBederKlicic2024}. Four future-oriented scenarios were developed into 16 BMC variants (Lean/Non-Profit, practical/visionary), expanding the strategic solution space before consolidating robust configurations~\citep{RothBederKlicic2026Brand}.
The Minimum Viable Product (MVP), implemented as a Figma prototype with core modules (matchmaking, analytics, knowledge hub), underwent staged validation: expert evaluation (N = 10), surveys (N = 92), and MVP testing (N = 5), assessing usability, governance fit, and strategic coherence~\citep{Kubullek2025}.
Together, these components form an integrated innovation system, where the RimA Toolbox provides methodological guidance and flexibility, transforming established models contextually to create a scalable, sustainable ecosystem for social robotics adoption.

\section{Discussion of Lessons Learned}
\label{sec:lessons_learned}
Interdisciplinary benchmarking was challenging, as stakeholders from different backgrounds often pursued different goals and prioritized different forms of expertise within the benchmarking process.
Interdisciplinarity, however, has been shown to be crucial for advancing domains from research to application~\citep{waldman2013interdisciplinary}.

As the related work (\cref{sec:relatedwork}) showed, no universally applicable benchmarking methods currently exist. The most common approach, comparing solutions outside the own laboratory in robotics competitions, is limited in the TRLs it can represent. Our benchmarking thus measured against panel-defined ex-ante criteria rather than an externally accepted standard (\cref{sec:relatedwork})---a limitation of the field itself that the process is designed to reduce.
As a result, benchmarking the interplay and potential tensions between task fulfillment, interaction quality, economic viability, and safety proved challenging. At the same time, the process itself provided value for the individual use cases by systematically revealing context-specific requirements, trade-offs, and deployment constraints even where direct cross-case comparability remained limited.

This is not per se a problem (yet), as few service robots are ready for everyday use and available to buy~\citep{ronnau2023towards}. The robots used in the three project-based use cases were Pepper, Angsa, and Adlatus CR700 and SR1300. These robots are products buyable in Germany but were used in (varying degrees of) prototypical application scenarios. 

Benchmarking the task fulfillment reveals a persistent challenge from robotics competitions: many solutions perform well in controlled laboratory settings but struggle in more realistic environments. This issue is not uniformly addressed by all benchmarked robots in the current approach. For instance, the \emph{Public Library} use case demonstrated comparable task fulfillment, whereas the \emph{Park Cleaning} scenario—conducted under realistic conditions in an actual park—encountered difficulties such as bottle caps embedded in grass. 
These findings illustrate how increasing environmental realism introduces context-specific challenges that are difficult to capture through uniform benchmarking schemes.

Unifying task fulfillment metrics across diverse use cases, particularly when their foci differ, remains impractical. The nature and frequency of interactions varied between Phases 2 and 3, leading to distinct impacts on task fulfillment. While functional metrics can be objectively measured or derived, interaction-centric metrics—such as dialog success rates—require more systematic evaluation. Additionally, disruptive factors like prolonged processing times or repetitive misunderstandings should be more explicitly reflected in these metrics.
A general limitation of the benchmarking approach is the limited sample rate, as emphasis lay on comparative cross-case analysis and the derivation of objective metrics.

For interaction quality, the evaluation highlighted structural limitations beyond individual use cases. The comprehensive guideline developed, covering a broad spectrum of interaction-related dimensions, proved too generic and extensive to serve as a standardized instrument across scenarios without substantial adaptation. Its main strength lies in its potential as a modular framework tailored to specific use cases rather than a universal benchmarking tool. The assessment of interaction quality remained predominantly subjective, as human–robot interaction is shaped by individual perception and context. Although first methodological approaches to objectifying HRI have been proposed (see \cref{sec:relatedwork-interaction}), they are not yet sufficiently established, validated, or readily applicable for fast, resource-constrained benchmarking settings; given the limited time frames of the events, their systematic integration was not feasible. 

Regarding the benchmarking of safety, there is some common ground: all robots need to avoid hurting or endangering the humans they interact with. In Germany, no safety standard applies to public everyday life interactions between robots and humans. Existing test objects simulate aspects like the legs of grown-ups, but not children, walking aids like poles, or objects that are difficult to detect for common sensors, like the leg of a grill. The benchmarking showed that both deployers and the panel struggled to benchmark safety beyond stairs and obstacle detection, as became clear in the outdoor scenarios.
Overall, benchmarking safety was stopped after Phase 2, as real-scenario benchmarking was out of scope.

Regarding economic viability, the results highlight that economic considerations tend to be addressed only implicitly or late in the development of service robotics applications and strategic networks, despite their central relevance for real-world, long-term deployment. A recurring challenge identified across cases was the tension between designing solutions that are useful and inclusive while at the same time preventing costs from escalating beyond sustainable levels. The benchmarking further revealed substantial differences in analytical depth, largely driven by varying levels of prior knowledge among participants, underscoring the need for structured, phase-specific guidance that explicitly integrates economic viability from the early phases onward.

The overall lesson from the three-phase experiment was that the interaction of the categories (task fulfillment, interaction quality, safety, and economic viability) becomes visible through this approach -- both in its tension and its complexity. From a meta-level perspective, economic viability emerged as a cross-cutting dimension that connects micro-level use case design, meso-level sustainability of a concept, and macro-level considerations of scalability and platformization. The benchmarking process made visible that neglecting one of these levels risks local optimization at the expense of long-term impact. Consequently, neither of the categories should be understood as a single evaluative criterion but as a multi-level design challenge within interdisciplinary benchmarking.

Regarding the goal of interdisciplinary benchmarking of three TRL 6-7 everyday life scenarios \cite{ronnau2023towards}, the use cases and the panel faced the organizational limitations of the approach. While short-duration tests were conducted in real environments in the presence of the benchmarking panel, long-term tests proved out of scope. 

The present study already used use-case-specific metrics. The next step is therefore to make their selection and documentation more systematic and reproducible. A revised modular framework would retain the four coordinated evaluation categories and three phases while requiring a common use-case profile and explicit justification of each metric’s relevance, operationalization, applicability, and comparability. Changes and exclusions should be documented, and direct quantitative comparisons should be limited to sufficiently similar cases. Safety remains a mandatory evaluation dimension and cannot be offset by performance in other categories. Generalizability would therefore lie in the shared procedure rather than in identical metrics.


\section{Conclusion}
\label{sec:conclusion}
 Advancing robots from lab research to deployment in practice is a current challenge in  the research area of benchmarking  robots.   This paper investigated how service robots can be benchmarked under realistic deployment conditions at TRL 6–7. Rather than relying solely on standardized laboratory comparisons, the proposed process evaluated robots across multiple public real-world scenarios and complementary evaluation dimensions with an interdisciplinary benchmarking panel.

Overall, the chosen benchmarking process did not yield directly comparable quantitative results across the three heterogeneous use cases. However, it generated valuable meta-level insights into interdisciplinary benchmarking -- especially regarding the interaction of task fulfillment with solutions for interaction quality, safety, and economic viability.  From an exploratory and qualitative perspective, stepping into real-world practice outside a competitive setting—following the EuRoC idea—proved to be a highly insightful alternative to traditional robotics competition formats.

This approach successfully highlighted challenges that are inherently interdisciplinary and not confined to a specific use case or robotic platform. While it required considerable travel and time investment, it produced meaningful, experience-based learnings, particularly for the benchmarking panel experts, and more broadly for researchers seeking to benchmark and compare their robots within specific application contexts. 


The findings suggest that future benchmarking frameworks should retain task fulfillment, interaction quality, safety, and economic viability as distinct but coordinated evaluation modules within a common selection and reporting process. The framework should not prescribe a fixed set of metrics. Instead, a shared use-case profile should guide metric selection, and the relevance, operationalization, applicability, and comparability of each metric should be documented. This approach enables process-level comparability across heterogeneous deployments while limiting direct quantitative comparisons to sufficiently similar cases. Safety remains a mandatory requirement that cannot be offset by performance in other categories. Benchmarking at higher TRLs would thus provide structured evidence about deployment-specific performance and trade-offs rather than a universal score. As this proposal is derived from three use cases, it requires validation in further deployments.

\section{Ethics Statement} 
Ethical approval was not required for the studies involving humans because the work focused exclusively on benchmarking the performance of robotic systems and did not involve human participants as research subjects. No personal, identifiable, or sensitive human data were collected, recorded, or analyzed. The evaluation was performed by an expert panel solely for the purpose of assessing robot performance against predefined benchmark criteria. All experimental activities were conducted in controlled field environments and in accordance with established safety procedures. The study involved only low-risk, non-invasive observation of robot behavior and did not include interventions, measurements, or data collection involving human participants. The studies were conducted in accordance with the local legislation and institutional requirements. Written informed consent for participation was not required from the participants or the participants' legal guardians/next of kin in accordance with the national legislation and institutional requirements because as no individuals participated as research subjects and no identifiable personal data were collected. Photographs included in this manuscript are provided solely to illustrate the experimental setup. Any individuals who may appear incidentally in the images have been anonymized, and no identifiable personal information is disclosed.

\section{Acknowledgments}
This work has been funded by the German Ministry of Research, Technology and Space
(BMFTR), grant nos. 16SV8680, 16SV8681, 16SV8683, 16SV9263, 16SV8941, 16SV8693 as well as 16SV8934 and was headed out of the project \textquote{Transferzentrum Roboter im Alltag (RimA)}.

The planning and execution of the benchmarking events involved numerous researchers from numerous institutes. We thank all our colleagues from the projects \emph{RimA}, \emph{rokit}, \emph{RuhrBots}, and \emph{ZEN-MRI} for their contributions and willingness to experiment with and to discuss this setup.

\bibliographystyle{Frontiers-Harvard}
\bibliography{references}

\end{document}